%% file: main.tex
\documentclass[12pt]{article}

\usepackage{fancyhdr}       %
\usepackage[utf8]{inputenc} %
\usepackage[T1]{fontenc}    %
\usepackage{hyperref}       %
\usepackage{natbib}
\usepackage{fullpage}
\usepackage{url}            %
\usepackage{booktabs}       %
\usepackage{amsfonts}       %
\usepackage{nicefrac}       %
\usepackage{microtype}      %
\usepackage{xcolor}         %
\usepackage{amsmath}
\usepackage{graphicx} 
\usepackage[parfill]{parskip}

\usepackage{soul}
\usepackage[normalem]{ulem} 

\input{preamble}

\title{Operator-informed score matching for Markov diffusion models}

\author{Zheyang Shen$^1$, Huihui Wang$^2$, Marina Riabiz$^2$, Chris. J. Oates$^{1,3}$ \\
  $^1$Newcastle University, UK \\
  $^2$King's College London, UK \\
  $^3$The Alan Turing Institute, UK 
}
\date{}

\begin{document}
\maketitle
\begin{abstract}

\input{sections/abstract}
\end{abstract}

\input{sections/introduction}

\input{sections/background}

\input{sections/main_discussion}

\input{sections/experiments}
\input{sections/post_discussion}

\paragraph{Acknowledgments}
\input{sections/ack}

\bibliographystyle{abbrvnat}
\bibliography{references}

\newpage
\appendix

\input{sections/supplement_ve}
\FloatBarrier

\end{document}

%% file: sections/abstract.tex
Diffusion models are typically trained using score matching, a learning objective agnostic to the underlying noising process that guides the model. 
%This paper argues that \emph{Markov} noising processes enjoy an advantage over generic noising frameworks, as their associated Markov operators are better understood.
This paper argues that \emph{Markov} noising processes enjoy an advantage over alternatives, as the Markov operators that govern the noising process are well-understood.
Specifically, by leveraging the spectral decomposition of the infinitesimal generator of the Markov noising process, we obtain parametric estimates of the score functions simultaneously for all marginal distributions, using only sample averages with respect to the data distribution.
The resulting \emph{operator-informed score matching} provides both a standalone approach to sample generation for low-dimensional distributions, as well as a recipe for better informed neural score estimators in high-dimensional settings.

%% file: sections/introduction.tex
\Acp{dm} \citep{sohl-dicksteinDeepUnsupervisedLearning2015,songGenerativeModelingEstimating2019,hoDenoisingDiffusionProbabilistic2020} %\textcolor{red}{\sout{have demonstrated remarkable performance in generative modeling, including in many instances where data are complicated and high-dimensional.}
have emerged as powerful generative models, achieving strong performance on complex, high-dimensional data.% %\mr{Do we mention any application?} 
Their conceptual appeal stems from the simple yet illuminating observation that ``creating noise from data is easy''~\citep{songScoreBasedGenerativeModeling2020}, as a wide range of %\textcolor{red}{\sout{myriad} a wide range of} 
tractable diffusion processes (often known as \emph{forward} or \emph{noising processes}) can be used to devolve a hard-to-sample data distribution $p_{\text{data}}$ into an easy-to-sample noise distribution $\pi$. 
The forward process induces a corresponding \emph{backward process}, whose law coincides with the time reversal of the forward process, which can be simulated provided that the score functions of its marginal distributions are known. 
As simulating the backward process produces samples from $p_{\text{data}}$, much effort has been devoted to \emph{score matching}, a data-driven approach that infers the score function from independent samples \citep{hyvarinenEstimationNonNormalizedStatistical2005, vincentConnectionScoreMatching2011,songSlicedScoreMatching2020}.
%and can be approximated using techniques such as score matching \citep{hyvarinenEstimationNonNormalizedStatistical2005, vincentConnectionScoreMatching2011}.
%Simulation of the backward process enables \textcolor{red}{\sout{ samples from $p_{\text{data}}$ to be approximately generated.} to approximately generate samples from $p_{\text{data}}$.}

%Markov \acp{dm} in particular have been shown to provide state-of-the-art performance on tasks such as image generation \citep{rombach2022high}, text-to-3D \citep{poole2022dreamfusion}, and prediction of molecular binding structure \citep{corso2023diffdock}.

While Markov \acp{dm} remain the default choice for forward processes, they play a supporting role in degrading structured data into unstructured noise. In principle, a wide variety of forward processes can be reversed for generative modeling;
%there exists considerable flexibility in how noise is incorporated within a \ac{dm}; 
for example, it is not required that the forward process is Markov, or even a diffusion process (despite the \ac{dm} terminology) \citep{bansalColdDiffusionInverting2023}. 
Although the generality of the \ac{dm} framework exceeds the breadth of \emph{Markov} forward processes, we illustrate that a careful analysis of Markov forward processes confers the following new insight and methodology:
%Regardless of how the forward process is specified, accurate approximation of the noise-perturbed score functions is critical, and this is typically achieved using generic estimation techniques, such as score matching applied to a sufficiently large dataset. 
%However, while all time-indexed sequences of noise-perturbed distributions can be inverted given accurate approximations to their score functions, the aim of this paper is to illustrate that \emph{Markov} forward processes offer important additional new insight and functionality for a \ac{dm}:
\begin{itemize}[leftmargin=*]
    %\item \textbf{Exact solution of the forward process:} A Markov forward process has a formal solution as a sequence of time-dependent \acp{kme} of the data-generating distribution \citep{smolaHilbertSpaceEmbedding2007}, meaning that in principle both forward and backward processes can be \emph{exactly solved}.  
    %Though in practice the data-generating distribution is not directly observed, this insight enables the development of a novel technique called \emph{Riemannian diffusion kernel smoothing}, which can ameliorate the burden of neural score approximation for \ac{dm}, at least in a low-dimensional context.
    \item \textbf{New insight into score matching:} 
    The evolution of the data distribution under the forward process is best understood from the perspective of \emph{Markov diffusion operators} -- an operator-theoretic study of conditional expectations. 
    %\textcolor{teal}{[ZS: I don't know the correct punctuation there]}
    In particular, the most popular Markov forward processes have well-defined diffusion operators that can be explicitly computed, which we leverage to derive an operator-based variant of the score matching loss. 
    %Self-adjointness of diffusion operators streamlines the derivation of tractable formula for score matching, and by extension, the $L_2$ minimization of the conditional expectation of any functional. 
    %
    %
    %Streamlined derivations for time score matching \citep{choiDensityRatioEstimation2022} and the score Fokker--Planck equation \citep{laiImprovingScorebasedDiffusion2023} can also be obtained.
    \item \textbf{Improved score matching:} The \emph{spectrum} of diffusion operators characterizes the temporal evolution of conditional expectations, and for the most popular Markov forward processes the spectra can be explicitly computed. 
    Drawing parallels from \emph{orthogonal function} methods, we propose \emph{operator-informed score matching} (\acs{oism}); a novel, widely applicable
    %, and straight-forward to implement 
    approach to score matching that exploits an explicitly computable approximation to the score function based on the eigenfunctions of the infinitesimal generator. 
    Our approach is notable for its simplicity, as we require only estimates of the expected value of eigenfunctions with respect to the data distribution to produce score approximations across all noise perturbation levels. We showcase the validity of our approach with illustrative and practical examples, which demonstrate that \ac{oism} is expressive enough on its own to generate low-dimensional data distributions, and the simple addition of low-order \ac{oism} approximations accelerates the training of neural network score estimates. 
    %This has the effect of reducing sample complexity, and is empirically demonstrated to improve performance of the trained generative model.
    %\textcolor{blue}{[CJO:  or should we say, `to accelerate training' instead (or both)?]}
\end{itemize}
%\textcolor{teal}{[ZS: I changed the above section -- mostly it is now a bit wordy as diffusion operators were not introduced until this point. ]}\mr{I also left going through this section for later.}

%
%
%
%

% \textcolor{red}{\sout{.  Markov diffusion operators present}. This formalism provides} 
% \textcolor{red}{\sout{The present paper focuses on the theoretical foundations of Markov \acp{dm} and leverages these insights to develop improved score matching methodology.To achieve this, we adopt} } 
The interplay between score matching methodology and the choice of the forward process is a relatively unexplored research direction in \acp{dm}.
The recent works of \citet{scarvelisClosedFormDiffusionModels2023,laiImprovingScorebasedDiffusion2023,bentonDenoisingDiffusionsDenoising2024} address this aspect, but to a lesser extent compared to the present paper, as discussed in detail in \Cref{sec: discuss}.
Our work adopts the framework of \emph{Markov diffusion operators} as set out in standard references such as \citet{bakryAnalysisGeometryMarkov2013}.
This formalism provides an aerial view of the corresponding forward process as an evolution of probability guided by Markov semi-group operators, and we contend that this perspective enables useful theoretical insights, as well as the development of improved score matching methodology, in the context of \acp{dm}.

The set-up and notation used in this paper is introduced in \Cref{sec: set-up}.
Our main findings on \acp{dm}, obtained through the lens of Markov diffusion operators, are detailed in \Cref{sec: lens}.
%Consequences of an explicit closed form solution to the forward process are explored in \Cref{sec: exact solution}, while 
The proposed \ac{oism} methodology is presented in \Cref{sec: operator informed} and empirically assessed in \Cref{sec: empirical}.
An extended discussion is contained in \Cref{sec: discuss}.

%% file: sections/background.tex
\section{Set-up and Notation}
\label{sec: set-up}

Simply put, the central task of \emph{generative modeling} is learning to produce novel samples from a data distribution $p_{\text{data}}$, given only a finite set $\{\mbx_m\}_{m=1}^M$ of independent samples from $p_{\text{data}}$. 
% \mr{I am not sure if we should use $p_{\text{data}}$ at all, or $\rho_0$ from the start. But I think that after introducing $\rho_0$ we should stick with it, eg in equations 1 and 2. The notation used for the data was also ambiguous with respect to the  notation that denotes realization of the DM, but I have changed the latter from $\mbx_0$ to $\mbx$, I hope this is ok.}
% \textcolor{teal}{[ZS: I don't mind both $p_{\text{data}}$ and $\rho_0$ coming up in the text -- I think one is to emphasize it comes from the data distribution, the other is to simplify notation when we analyze marginal distributions. ]}
%Given a set $\{\mbx_n\}_{n=1}^N$ of independent samples from $p_{\text{data}}$, the task is to construct a \emph{generative model}, meaning an accurate approximation to $p_{\text{data}}$ that can be easily sampled.
For concreteness, we presume in this paper that $p_{\text{data}}$ is supported on $\mathbb{R}^d$, and for convenience, we overload the notation so that (e.g.,) $p_{\text{data}}$ refers both to the distribution and its associated density on~$\mathbb{R}^d$.
Similarly, we let $\mathcal{N}(\mbmu,\mbSigma)$ denote the Gaussian distribution with mean $\mbmu \in \mathbb{R}^d$ and covariance $\mbSigma \in \mathbb{R}^{d \times d}$, and $\mathcal{N}(\cdot \vert \mbmu,\mbSigma)$ denote the corresponding density function in shorthand. 
%\textcolor{red}{$\mbX$ denotes a random vector, and  $\mbx$ its realization.} \mr{Do we want to move here the full notation about norms, gradients etc.?}\mr{$\mathrm{d}\mbx$ denotes the Lebesgue measure.}
For the purpose of this paper, we assume all relevant densities exist and are positive and differentiable as required, and that all integrals we consider exist.

Here we briefly summarize the key elements of generative models based on \acp{dm}.
Our aim is not to provide a rigorous mathematical treatment, as comprehensive introductions to the stochastic analysis tools underlying \acp{dm} can be found in \citet{oksendalStochasticDifferentialEquations2003,stroock2013introduction,sarkkaAppliedStochasticDifferential2019,bakryAnalysisGeometryMarkov2013}.
The \emph{forward process} of a \ac{dm} is a time-indexed stochastic process $(\mbX_t)_{t \geq 0}$ whose initial value $\mbX_0 \in \mathbb{R}^d$ is sampled from $\rho_0 := p_{\text{data}}$, the distribution which gives rise to the dataset, and for which the marginal distributions $\left(\rho_t\right)_{t\geq 0}$ of subsequent states $\mbX_t$ converge to an explicitly known noise distribution $\pi$ as $t \rightarrow \infty$.
Prominent examples of Markov \acp{dm} include the \emph{variance-exploding}  \ac{dm}
%\textcolor{teal}{[ZS: I understand the $\sqrt{2}$ below is a bit confusing -- it is mainly added for consistency, otherwise $\Gamma(f, g) = \frac{\langle\nabla f, \nabla g\rangle}{2}$.]}\mr{The $\sqrt{2}$ here is ok, and consistent with that in the OU process.}
\begin{align}
    \d{\mbX_t} = \sqrt{2}\d{\mbW_t}, \qquad \mbX_0 \sim 
        \rho_{0}, 
    % p_{\text{data}}, 
    \label{eq: bm process}
\end{align}
where $\mbW_t$ is a standard $d$-dimensional \ac{bm}, and the \emph{variance-preserving} \ac{dm}
\begin{align}
    \d{\mbX_t} = -\mbX_t\d{t} + \sqrt{2}\d{\mbW_t}, \qquad \mbX_0 \sim 
    \rho_{0}, 
    % p_{\text{data}}, 
    \label{eq: OU process}
\end{align}
also known as an \ac{ou} process \citep{songScoreBasedGenerativeModeling2020}.
For the \ac{ou} process, it is well-known that the marginal distribution $\rho_t$ converges to the standard Gaussian distribution $\pi = \cN(\mathbf{0}, \mbI)$ as $t\rightarrow \infty$, and that the conditional distribution $\rho_t(\mbX_t\vert \mbX_0 =\mbx)$, %\textcolor{red}{\sout{conditioned on the initial point $\mbx$,} 
given that $\mbX_0$ takes value $\mbx$, coincides with $\cN(\mbX_t\vert \alpha_t\mbx, \sigma_t^2\mbI)$, where $\alpha_t := e^{-t}$, and $\sigma_t^2 := 1-\alpha_t^2$. %\mr{Do we need to give invariant and conditional distributions for the VE case?}
In practice, the forward process is simulated until a finite time horizon $T$, large enough that $\rho_T$ becomes indistinguishable from $\pi$. %\mr{Need to define what $\approx$ means}
The \emph{probability flow} \ac{ode}~\citep{songScoreBasedGenerativeModeling2020} 
\begin{align}
    \frac{\d{\mbX_t}}{\d{t}}= -\nabla\log \frac{\rho_t}{\pi}(\mbX_t) = -\mbX_t - \nabla\log \rho_t(\mbX_t), \qquad \mbX_0\sim 
    % p_{\text{data}}
    \rho_0
    \label{eq: prob flow ode}
\end{align}
describes the evolution of the marginal distributions $\rho_t$ under the \ac{ou} dynamics. 
%\mr{Add initial condition. As written here, the probability flow generates samples, which is indeed the main goal. We can introduce evolution of $\rho_t$ via probability flow in the supplement for evaluation. Do we want to mention also the reverse SDE for sample generation? Do we need to present VE as well?}
%\textcolor{teal}{[I think prob flow ODE described above is still in the forward direction, in order to underscore the similarity in $\rho_t$. SDE sampler is a technicality from my perspective.]}
The main observation in \ac{dm} is that simulating \eqref{eq: prob flow ode} backwards in time (the \emph{backward process}) with initialization $\mbX_T \sim \pi$ approximately yields $\mbX_0\sim\rho_0$. 
To instantiate the backward process, the score of $\rho_t$, that is, $\nabla\log \rho_t(\mbx)$, is required. 
%\textcolor{teal}{[ZS: later in the text we provide foreshadowing about data-driven score estimation, so we can keep this explanation as short as possible.]}
%\textcolor{red}{and}  \citep{hyvarinenEstimationNonNormalizedStatistical2005,vincentConnectionScoreMatching2011} \textcolor{red}{provide widely used score matching methodology}. \mr{Introduce DSM vs ISM here? Mention here that this provides a loss function, optimized via NN?  Emphasize that forward samples are used for the training, and perhaps contrast already with OISM.} 

Markov forward processes remain the default choice for~\acp{dm}, with the \ac{bm} process in \eqref{eq: bm process} and the \ac{ou} process in \eqref{eq: OU process} most widely used.
However, there are many other forward processes yielding a time-indexed sequence of distributions of decreasing complexity that can in principle be inverted~\citep[e.g.][]{bansalColdDiffusionInverting2023}. 
This paper presents new theoretical insights that highlight particular advantages of \emph{Markov} forward processes, relative to these alternatives, in the context of \acp{dm}.
To this end, we next revisit \acp{dm} through the lens of Markov diffusion operators.

%% file: sections/main_discussion.tex
\section{Diffusion Models through the Lens of Markov Diffusion Operators}
\label{sec: lens}

This section outlines the basis for our theoretical insight into Markov \acp{dm}.
Our main tools are Markov diffusion operators \citep{bakryAnalysisGeometryMarkov2013}, a set of operators that can be used to characterize the marginal distributions $(\rho_t)_{t \geq 0}$ of a Markov \ac{dm} (\Cref{subsec: spectral properties}).
These operators will be 
%demonstrated to enable exact solution of the forward process (\Cref{subsec: exact solution}), and 
exploited to shed new light on score matching and related objectives (\Cref{subsec: rederive score matching}). 
%\mr{It might be nice to make an analogy between operator and matrices somewhere, to give intuition to things like the exponential of an operator, the spectrum etc.} \textcolor{teal}{[ZS: I don't mind skipping over the matrix analogy, as we now have the addendum in explaining the exponential. ]}

%

%

%\subsection{Spectral Properties of Markov Diffusion Forward Processes}
\subsection{Markov Diffusion Operators and their Spectral Decompositions}
\label{subsec: spectral properties}
One of the fundamental ideas that we leverage in this paper is that the convergence of a Markov diffusion process to its stationary distribution $\pi$ can be characterized using its Markov semigroup and infinitesimal generator, as explained below. 
The \emph{Markov semigroup} $(P_t)_{t\geq 0}$ is a time-indexed sequence of operators, such that $P_t f(\mbx)$ evaluates the conditional expected value of $f(\mbX_t)$ given that $\mbX_0$ is initialized at $\mbx$, that is, 
\begin{align*}
    P_t f(\mbx) := \mathbb{E}_{\mbX_t \sim \rho_t\left(\mbX_t\vert\mbX_0=\mbx\right)} \left[f(\mbX_t)\right] . %
\end{align*}
From this definition, and using the tower rule, it is easy to derive the following identity, which we will use throughout
\begin{align}
\mathbb{E}_{\mbX_t \sim \rho_t} [ f(\mbX_t) ] = \mathbb{E}_{\mbX_0 \sim \rho_0} [ P_t f(\mbX_0) ].
\label{eq:evolution_expectations}
\end{align}
Moreover, we can represent $(P_t)_{t\geq 0}$ more parsimoniously using its \emph{infinitesimal generator} $\mathscr{L}$, which is defined as the derivative of $P_t$ at $t=0$:
\begin{align*}
    \mathscr{L} f(\mbx) := \lim_{t\downarrow 0} \frac{P_t f(\mbx) - f(\mbx)}{t},
\end{align*}
noticing that $P_0f(\mbx) = f(\mbx)$.
As a result, we can formally express $P_t = e^{t\mathscr{L}}$, where the exponential of an operator is defined by the Taylor expansion of the exponential function, in analogy to the matrix exponential. 
% For a $\pi$-invariant Markov diffusion process, $\mathbb{E}_{\mbX \sim \pi} [P_t f(\mbX)]$ is equal to $\mathbb{E}_{\mbX \sim \pi} [f(\mbX)]$ for all $t$, and
% %Since a stationary measure $\pi$ of a Markov diffusion processes satisfies $\mathbb{E}_{\mbX \sim \pi} [P_t f(\mbX)] = \mathbb{E}_{\mbX \sim \pi} [f(\mbX)]$, 
% it follows that the range of $\mathscr{L}$ contains only functions whose expectations are zero with respect to $\pi$. 
% \mr{Do we use this observation? We would need to define the range of an operator. Instead here we could say that it holds that 
% \begin{align}
% \mathbb{E}_{\mbX_t \sim \rho_t} [ f(\mbX_t) ] = \mathbb{E}_{\mbX_0 \sim \rho_0} [ P_t f(\mbX_0) ].
% \label{eq:evolution_expectations}
% \end{align}
% It might also be a good place where to define the 
% adjoint operator and state $\rho_t = P_t^*\rho_0$, 
% and the Fokker Plank $\partial_t \rho_t = \mathcal{L}^* \rho_t$, with respect to the Lebesgue measure. However, operators are symmetric with respect to $L^2(\pi)$ and, as a consequence, 
% $\partial_t \frac{\rho_t}{\pi} = \mathcal{L} \frac{\rho_t}{\pi}. $
% This is also part of the reason why we work with density ratios. This would more formally deliver what we promised to say in the introduction - how marginal distributions evolve via operators.} 

% \mr{If we need more space, the conditional distributions for the Markov semigroups were defined in Section 1, and we could make a table for the infinitesimal generators and their spectrum. }
\begin{example}[\ac{bm} process] \label{ex: bm}
    The Markov semigroup associated with the \ac{bm} process in \eqref{eq: bm process} is $P_t f(\mbx) = \mathbb{E}_{\mbX_t \sim \mathcal{N}(\alpha_t \mbx, \sigma_t^2 \mbI)} \left[f(\mbX_t)\right]$, where $\alpha_t = 1$ and $\sigma_t^2 = 2t$ and the infinitesimal generator is $\mathscr{L} f(\mbx) = \Delta f(\mbx)$; see Example 7.3.4 of \citet{oksendalStochasticDifferentialEquations2003}. 
    The invariant measure of \ac{bm}, while no longer a probability measure, is nominally the Lebesgue measure $\mathrm{d}\mbx$ given its infinite variance as $t\rightarrow\infty$. 
\end{example}

\begin{example}[\ac{ou} process] \label{ex: ou}
    The Markov semigroup associated with the \ac{ou} process in \eqref{eq: OU process} is $P_t f(\mbx) = \mathbb{E}_{\mbX_t \sim \mathcal{N}(\alpha_t\mbx, \sigma^2_t\mbI)} \left[f(\mbX_t)\right]$, where $\alpha_t = e^{-t}$, $\sigma^2_t = 1-\alpha_t^2$, and the infinitesimal generator is $\mathscr{L} f(\mbx) = \langle -\mbx, \nabla f(\mbx) \rangle + \Delta f(\mbx)$; see Theorem 7.3.3 of \citet{oksendalStochasticDifferentialEquations2003}.
\end{example}

Further examples include, e.g., the critically-dampened \ac{ou} process \citep{dockhornScoreBasedGenerativeModeling2021}.
For brevity, in the sequel we focus on the \ac{ou} forward process, 
but our derivations can be reproduced for other Markov diffusion processes.
A formulation for Markov forward processes corresponding to general \ac{lti} \ac{sde}~\citep{sarkkaAppliedStochasticDifferential2019} is given in \Cref{app: brownian}.

In summary, the infinitesimal generator $\mathscr{L}$ characterizes how conditional expectations evolve over time. While it is generally nontrivial to explore the spectral properties of an arbitrary diffusion process \citep{chewiSVGDKernelizedWasserstein2020}, the spectra of Markov diffusion processes are well-understood \citep{bakryAnalysisGeometryMarkov2013}. 
The \emph{spectrum} of a Markov diffusion processes consists of pairs of eigenvalues $\lambda_n$ and eigenfunctions $\phi_n$, such that $\mathscr{L} \phi_n(\mbx) = \lambda_n\phi_n(\mbx)$, and will be denoted $\left(\lambda_n, \phi_n\right)_{n\in\mathcal{I}}$ for an appropriate index set $\mathcal{I}$.
Because $P_t$ is formally $e^{t\mathscr{L}}$, we have $P_t\phi_n=e^{\lambda_n t}\phi_n$ and the spectrum of the semi-group operator $P_t$ is $(e^{\lambda_n t}, \phi_n)_{n \in \mathcal{I}}$. Therefore
%Armed with the concept of an infinitesimal generator, we can now define the \emph{spectrum} of a Markov diffusion process as consisting of eigenpairs $\left(\lambda_n, \phi_n\right)_{n\geq 0}$ of the infinitesimal generator, such that $\mathscr{L} \phi_n(\mbx) = \lambda_n\phi_n(\mbx)$. 
%The spectrum of the Markov semigroup can be similarly defined as consisting of pairs $(e^{\lambda_n t}, \phi_n)_{n \geq 0}$, which satisfy the eigen-relation
\begin{align}
    \mathbb{E}_{\mbX_t \sim \rho_t} [ \phi_n(\mbX_t) ] &= \mathbb{E}_{\mbX_0 \sim \rho_0} [ P_t \phi_n(\mbX_0) ] = e^{\lambda_n t} \mathbb{E}_{\mbX_0 \sim \rho_0} [ \phi_n(\mbX_0) ] , \label{eq: eigenrelation}
\end{align}
where the first identity holds because of \eqref{eq:evolution_expectations}. This implies that that the expectations of the eigenfunctions  $\phi_n$ 
% \textcolor{red}{with respect to the time marginals $\rho_t$} 
\emph{evolve predictably}  under the dynamics of the forward process as governed by the exponent $\lambda_n$ in \eqref{eq: eigenrelation}.
This observation is key to the \ac{oism} methodology proposed in \Cref{sec: operator informed}: that is, the numerical approximation of quantities of the form of $\mathbb{E}_{\mbX_t\sim\rho_t}[f(\mbX_t)]$, a common occurrence in the training of \acp{dm}, can be bypassed for all $f$ constructed as linear combinations of eigenfunctions $\{\phi_n\}_{n \in \mathcal{I}}$. 
This begs the questions \emph{how expressive are these eigenfunctions}, and \emph{can an eigenfunction basis be harnessed to accelerate training of a Markov \ac{dm}?}
%\mr{Here we might want to carefully explain what the advantage will be: not just not needing samples from $\rho_t$, but having a parametric expression of the score, which is alternative, or helping the NN, reinforcing the message in the abstract.}
%one \emph{does not need to learn the time evolution of any score approximation that is based on eigenfunctions, since it is fully determined by the \ac{dm}}. \textcolor{teal}{[ZS: I find it hard to understand the above conclusion sentence. I wanted to come up with something like "the score function at time $t$ is fully determined by the initial distribution $\rho_0$, and informative score estimates might be attained if they are expressed in the coordinates of eigenfunctions", but that also sounds hard to understand. ]}

\begin{example}[\ac{bm} process, continued]
\label{ex: bm spectrum}
    The Laplace operator $\Delta$ coincides with the infinitesimal generator of the \ac{bm} process, and its eigenfunctions are the complex exponentials (that is, trigonometric functions) $\Delta \exp\left(\imath\langle \mbxi,\mbx\rangle\right) = -\norm{\mbxi}^2  \exp\left(\imath\langle \mbxi,\mbx\rangle\right)$, indexed by $\mbxi \in \mathbb{R}^d$. 
    Notice that in this case the spectrum is uncountable ($\mathcal{I} = \mathbb{R}^d$) and the eigenvalues are $\lambda_\xi = -\norm{\mbxi}^2$.
\end{example}

\begin{example}[\ac{ou} process, continued]
\label{ex: ou again}
    In the one-dimensional setting ($d=1$), the infinitesimal generator of the \ac{ou} process has a spectrum consisting of eigenpairs $(\lambda_n,\phi_n)_{n \in \mathbb{N}}$ with $\lambda_n = -n$ and $\phi_n(x) = \mathrm{He}_n(x) / \sqrt{n!}$, with $\mathrm{He}_n(x)$ the $n$th probabilist's Hermite polynomial \citep{abramowitz1965handbook}.
    In the multivariate setting ($d > 1$) the eigenpairs become tensorized as sums $\lambda_{n_1}+\cdots+\lambda_{n_d}$ and products $\phi_{n_1}(x_1) \cdots \phi_{n_d}(x_d)$ of the one-dimensional eigenpairs, indexed by $(n_1, \dots , n_d) \in \mathbb{N}^d$.
    % \textcolor{blue}{[CJO:  This previously said ``indexed by $\{(n_1, \dots , n_d) : n_1,\dots,n_d \geq 0 \}$''; hope this version is now correct?]}
    In each case, the eigenfunctions are a \emph{rich approximating set}; they form an orthogonal basis for $L^2(\pi)$ \citep[][Section 2.7.1]{bakryAnalysisGeometryMarkov2013}. 
\end{example}
%\textcolor{teal}{[ZS: is there a good reference to Hermite polynomials?]} \mr{\cite{abramowitz1965handbook} or its modern digital version \cite{NIST:DLMF} or yet\cite{szeg1939orthogonal}.}

\subsection{Score Matching via Generalized Integration-by-Parts}
\label{subsec: rederive score matching}
%\textcolor{teal}{[ZS: I'm in favor of mainly using $\Gamma$ instead of $\langle\nabla f, \nabla g\rangle$, to emphasize how score matching connects to the operator view.]}
The accurate approximation of the score function $\nabla\log \rho_t (\mbx)$ is the cornerstone to the success of a \ac{dm}, enabling samples to be generated by (e.g.) simulating the time reversal of the probability flow \ac{ode} \eqref{eq: prob flow ode}. 
The standard solution is to train a flexible time-dependent vector field $\mbs_t(\mbx)$ by minimizing the \emph{score matching loss}, an $L^2$ distance to the true score function:
%using score matching \citep{hyvarinenEstimationNonNormalizedStatistical2005}; i.e. to minimize an $L^2$ distance to the true score function $\mbs_t^\star(\mbx)$:
\begin{align}
    \min_{\mbs} \int_{t=0}^T \mathbb{E}_{\mbX_t \sim \rho_t} \left[\norm{\mbs_t(\mbX_t) - \nabla\log \rho_t (\mbX_t)}^2 \right] \; w(t) \; \mathrm{d}t, \label{eq: true score}
\end{align}
for some non-negative weight function $w(t)$. % that must be specified.
While \eqref{eq: true score} includes intractable terms, closed-form equivalent objectives can be obtained via \ac{ism} \citep{hyvarinenEstimationNonNormalizedStatistical2005}
or \ac{dsm} \citep{vincentConnectionScoreMatching2011}. 
Our first contribution in this paper is to note that the ISM loss function, as originally developed by \citet{hyvarinenEstimationNonNormalizedStatistical2005}, can be derived elegantly through the lens of Markov diffusion operators. 
% \mr{because of the many analogies we draw, it might be helpful to derive the classical ISM.}

%We note that when equipped with the family of Markov diffusion operators, the integration by parts formula originally developed by \citet{hyvarinenEstimationNonNormalizedStatistical2005} 
%the score matching objective \eqref{eq: true score} can be expressed more coherently. 
%Critically, from the point of view of this paper, standard practice is to learn the time dependence of the score function in a \emph{fully data-driven} manner, agnostic to the forward model.\nabla\log \rho_t (\mbX)
%Our operator-informed score matching methodology in \Cref{sec: operator informed} directly addresses this point, using knowledge of the forward model as in effect a variance reduction tool to lower the training burden of the \ac{dm}.
%However, before proceeding we first highlight another benefit of the Markov diffusion operator perspective; it enables a streamlined derivation of the standard score matching objective, as well as a new insight into related objectives, as will now be demonstrated.
%\textcolor{teal}{[There were some ``foreshadowing'' statements above, but it seems to me that we can foreshadow more effectively at the end of this section. What do you think? ]}

To streamline the following derivation, we introduce two bilinear operators induced by the infinitesimal generator $\mathscr{L}$, the \emph{carr\'{e}-du-champ} operator $\Gamma(\cdot, \cdot)$ and the \emph{Dirichlet} form $\cE(\cdot, \cdot)$, defined as: 
%The infinitesimal generator $\mathscr{L}$ naturally induces a general notion of integration by parts for integrals with respect to the stationary measure $\pi$ of the \ac{dm}, when given the bilinear \emph{carr\'{e}-du-champ} operator $\Gamma(\cdot, \cdot)$ and the \emph{Dirichlet} operator $\cE(\cdot, \cdot)$, defined as 
%To this end, we introduce the 
\begin{align*}
    \Gamma(f, g) := \frac{1}{2}\left(\mathscr{L}(fg) - f\mathscr{L}g - g\mathscr{L}f\right) , \quad \cE(f, g) := \int \Gamma(f(\mbx), g(\mbx))\d{\pi(\mbx)} ;
\end{align*}
see \citet{bakryAnalysisGeometryMarkov2013}, Section 1.4.2 for background. For \ac{bm} and \ac{ou} forward processes, their associated Markov semigroups are \emph{symmetric} on $L^2(\pi)$, yielding three equivalent forms of $\cE(f, g)$: 
% \mr{to state that this generalizes integration by part we only need two, consider removing one?}
%The inltegration by parts formula is then expressed as the self-adjointness of $\mathscr{L}$ when integrating over $\pi$; namely, 
\begin{align}
    \int \Gamma(f(\mbx), g(\mbx)) \d{\pi}(\mbx) = 
    % \cE(f(\mbx), g(\mbx)) = 
    -\int f(\mbx)\mathscr{L}g(\mbx)\d{\pi}(\mbx)=-\int g(\mbx)\mathscr{L}f(\mbx)\d{\pi(\mbx)} . \label{eq: gen int by parts}
\end{align}
Equation \eqref{eq: gen int by parts} can be seen as a generalization of the integration-by-parts formula with vanishing boundary term, which is recovered in the special case of the \ac{bm} process where $\mathscr{L}$ is the Laplace operator $\Delta$ and the invariant measure is $\mathrm{d}\pi = \mathrm{d}\mbx$.
%The symmetry of the Mehler kernel implies the self-adjointness of $\mathscr{L}$, see \citet[Chap. 1][]{bakryAnalysisGeometryMarkov2013}.
%The simple, familiar version of the integration by parts formula is recovered as the (degenerate) case corresponding to the \ac{bm} process, whose stationary measure is nominally the Lebesgue measure $\d{\mbx}$.
\begin{example}[\ac{ou} and \ac{bm} processes, continued]
\label{ex: operators for OU}
For the \ac{bm} and \ac{ou} processes, the carr\'{e}-du-champ operator is $\Gamma(f, g) = \langle \nabla f, \nabla g \rangle$ and the Dirichlet form is the $L^2(\pi)$ inner product $\cE(f,g) = \int \langle \nabla f (\mbx) , \nabla g(\mbx) \rangle \d{\pi}(\mbx)$, with respect to their invariant measures $\mathrm{d}\mbx$ or $\mathcal{N}(\mbx\vert\mathrm{0},\mbI)$.
\end{example}

There is an apparent semblance between the score matching loss \eqref{eq: true score} and the generalized integration-by-part formula defined using the carr\'{e}-du-champ operator, as they both involve the inner product of gradients. 
%\mr{This sentence is somewhat implicit, and there are few steps to get the second equation in \eqref{eq: midway score match}} 
% \textcolor{red}{\sout{This allows for a \emph{streamlined, $\pi$-aware} derivation of score matching.}}
%\paragraph{Streamlined derivation of score matching}
To see this, assume we estimate the density ratio $\frac{\rho_t}{\pi}$ as an \emph{energy-based model} $\frac{\exp(f_t(\mbx))}{Z_{f_t}}$ for some scalar-valued function $f_t$. Minimizing the expected squared distance between $\nabla f_t(\mbx)$ and $\nabla\log\frac{\rho_t}{\pi}(\mbx)$ can be seen as an equivalent, \emph{$\pi$-aware, streamlined} form of the traditional score matching loss \eqref{eq: true score}:
%Then the (un-weighted) integrand of the traditional score matching loss \eqref{eq: true score} can be rewritten in the equivalent 
%aims to minimize an $L^2$ loss between the score function and its approximation:
%\emph{$\pi$-aware, streamlined} form
\begin{align}
    \mathcal{L}(f_t) & := \mathbb{E}_{\mbX_t \sim \rho_t} \left[ \norm{\nabla f_t(\mbX_t) - \nabla\log \frac{\rho_t}{\pi}(\mbX_t)}^2 \right] \nonumber \\
    %& \stackrel{+C}{=} \mathbb{E}_{\mbX_t \sim \rho_t} \left[\norm{\nabla f_t(\mbX_t)}^2 - 2 \left\langle \nabla f_t(\mbX_t) , \nabla \log \frac{\rho_t}{\pi}(\mbX_t) \right\rangle \right] \label{eq: midway score match}
    & \stackrel{+C}{=} \mathbb{E}_{\mbX_t \sim \rho_t} \left[\Gamma(f_t(\mbX_t), f_t(\mbX_t)) - 2 \Gamma \left(f_t(\mbX_t) , \log \frac{\rho_t}{\pi}(\mbX_t) \right) \right], \label{eq: midway score match}
\end{align}
where the second equality features ``$+C$'' to indicate equality up to an $f_t$-independent additive constant.
The second term in \eqref{eq: midway score match} can be simplified using \eqref{eq: gen int by parts}, leading to
%simplified using (generalized) integration by parts \eqref{eq: gen int by parts}, to obtain the equivalent expression
\begin{align*}
    %\int \left\langle \nabla f_t(\mbx), \nabla \log \frac{\rho_t}{\pi}(\mbx) \right\rangle \d{\rho_t(\mbx)} & = \int \left\langle \nabla f_t(\mbx), \nabla \frac{\rho_t}{\pi}(\mbx) \right\rangle\d{\pi(\mbx)} \\
    %& \hspace{-30pt} = \cE\left(f_t, \frac{\rho_t}{\pi}\right) = -\int \mathscr{L}f_t(\mbx) \frac{\rho_t}{\pi}(\mbx)\d{\pi(\mbx)} = -\mathbb{E}_{\mbX_t\sim\rho_t} \mathscr{L}f_t(\mbX_t).
    \int \Gamma \left(f_t(\mbx) , \log \frac{\rho_t}{\pi}(\mbx) \right) \d{\rho_t(\mbx)} &= \int \Gamma \left(f_t(\mbx) , \frac{\rho_t}{\pi}(\mbx) \right) \d{\pi(\mbx)}\\
     & 
     %  \hspace{-5pt} 
     % = \cE\left(f_t, \frac{\rho_t}{\pi}\right) 
     = -\int \mathscr{L}f_t(\mbx) \frac{\rho_t}{\pi}(\mbx)\d{\pi(\mbx)} = -\mathbb{E}_{\mbX_t\sim\rho_t}\left[\mathscr{L}f_t(\mbX_t)\right].
\end{align*}
%\mr{Add a line about the step for the first equation} 
Substituting back into \eqref{eq: midway score match} and defining $\Gamma(f):=\Gamma(f, f)$, we arrive at a formula resembling the familiar \ac{ism} loss
% \textcolor{red}{[MR: Should it be $X_t$ in the integrand, not $X_0$?]}
\begin{align}
    \mathcal{L}(f_t) & \stackrel{+C}{=}
    \mathbb{E}_{\mbX_t \sim \rho_t} \left[\Gamma(f_t)(\mbX_t) + 2\mathscr{L} f_t(\mbX_t)\right]  \label{eq: hyvarinen score}
\end{align}
of \citet{hyvarinenEstimationNonNormalizedStatistical2005},
%, again in the special case of \ac{bm} forward process and its diffusion operators. 
% \mr{I have moved text around in this paragraph, I hope it conveys the same message you had in mind.}
which is equivalent to the un-weighted \ac{ism} integrand in \eqref{eq: true score} after the identification $\mbs_t(\mbx) = \nabla f_t(\mbx) + \nabla \log \pi(\mbx)$.

% \mr{do we want to to keep this sentence or say relation with Benton paper?}
The above derivation is novel, 
to the best of our knowledge, and we highlight that it is both streamlined and (somewhat) general.
However, the end result in \eqref{eq: hyvarinen score} is `just' the  \ac{ism} loss of \citet{hyvarinenEstimationNonNormalizedStatistical2005}.  
%In fact, if we subscribed to the conventional wisdom of training diffusion models in an entirely \emph{data-driven} manner, that is, producing samples $\mbX_t$ from noise-perturbed marginals $\rho_t$ and using them to calculate \eqref{eq: hyvarinen score} via Monte Carlo integration, whether or not we involved Markov diffusion operators in the derivation of a tractable score matching loss would make only notional difference. 
To tread new ground, we can make use of property \eqref{eq:evolution_expectations} of the semigroup $P_t$ and rewrite \eqref{eq: hyvarinen score} as an integral with respect to the noise-free data distribution~$\rho_0$, obtaining
%generalized to any infinitesimal generator $\mathscr{L}$, and requiring only regularity on $\rho_t/\pi$, instead of on the \emph{individual} densities of $\rho_t$ and $\pi$. \textcolor{teal}{[ZS: I think the above sentence makes sense after we introduce $\frac{\rho_t}{\pi} = P_t\frac{\rho_0}{\pi}$, so perhaps it's not yet needed. ]}
%As an additional bonus of the Markov diffusion operator perspective, we can exploit the symmetry of $P_t$ to shift the expectation with respect to the data distribution, so that
\begin{align}
    \mathcal{L}(f_t) & \stackrel{+C}{=} \mathbb{E}_{\mbX_0 \sim  \rho_0 } \left[
  P_t\Gamma(f_t)(\mbX_0) 
    + 2 P_t \mathscr{L} f_t(\mbX_0)\right]. \label{eq: score matching p0}
    %P_t\left[\norm{\nabla f_t(\mbX_0)}^2 + 2\mathscr{L} f_t(\mbX_0)\right] , \label{eq: score matching p0}
\end{align}
%which follows since $\mathbb{E}_{\mbX_t \sim\rho_t} g(\mbX_t) = \int g(\mbx) P_t \frac{\rho_0}{\pi}(\mbx)\d{\pi(\mbx)} = \mathbb{E}_{\mbX_0 \sim \rho_0} P_t g(\mbX_0)$.
%\textcolor{teal}{[ZS: I simplified the narrative above, as we don't need the symmetry to understand why it can be rewritten using $P_t$, because it is also true definitionally.]}
This last expression allows for a \emph{streamlined,  Markov-aware} and  \emph{operator-informed} derivation of score matching, 
% .
% However, the \emph{operator-informed} variant of score matching \eqref{eq: score matching p0} 
% \textcolor{red}{which} undergirds 
% key observations central to our methodology, \textcolor{red}{\sout{to be continued} that we build upon} in \Cref{sec: operator informed}\textcolor{red}{\sout{, namely}}: 
by providing two key observations for the methodology we develop in \Cref{sec: operator informed}: 
(i) the optimization of noised-perturbed scores   ($\nabla f_t$ in our re-parametrization)  links to expectations of quantities with respect to the data distribution $\rho_0$ via the operator $P_t$; (ii) sampling from the forward process is a general strategy to evaluate terms of the form $P_t \phi$, but when $\phi$ is linear combinations of eigenfunctions $\{\phi_n\}_{n \in \mathcal{I}}$, exact computation of $P_t \phi$ is possible using \eqref{eq: eigenrelation}, in principle enabling sampling from the forward process to be avoided. 
% \textcolor{teal}{[ZS: I have migrated most of the foreshadowing to this area. ]}

Markov diffusion operators also shed new light on other important aspects in diffusion modeling, 
%including \emph{Tweedie's formula} \citep{tweedieFunctionsStatisticalVariate1947}, 
including \emph{time score matching} \citep{choiDensityRatioEstimation2022}, a generalization of the \emph{score} \ac{fpe} \citep{laiImprovingScorebasedDiffusion2023}, and generalized score matching \citep{bentonDenoisingDiffusionsDenoising2024}. Since
these observations are not critical to our methodology, they are reserved for the interested reader in \Cref{app: consequences}.

\section{Operator-Informed Score Matching}
\label{sec: operator informed}

Building on the theoretical insights from the previous section, we present \emph{operator-informed score matching} (\acs{oism}), a recipe for score matching that makes use of the spectral decomposition of the Markov diffusion operators. At a high level, the concept of \ac{oism} is to approximate $\nabla \log (\rho_t/\pi)$ in the linear space spanned by eigenfunctions of the forward process, and leverage the spectral property \eqref{eq: eigenrelation} to compute both terms in \eqref{eq: score matching p0} in closed form, obtaining a quadratic form corresponding to the score matching loss. 
%Building on our theoretical insights up to this point, we now present a novel variance reduction technique called \acf{oism}, that is straightforward to implement within existing frameworks for \ac{dm}. 
%At a high level, the idea of \ac{oism} is to construct a first approximation to the score function based on eigenfunctions of the Markov \ac{dm}, and then to model the difference between the true and approximate scores using a neural network.
%Since the time-dependence of eigenfunctions is well-understood, constructing a first approximation is straight-forward and involves computing averages with respect to the original dataset.
%As a result there is little computational overhead associated with \ac{oism}, and yet the overall training burden for a \ac{dm} can be substantially reduced, as we show in \Cref{sec: empirical}.

\subsection{Score Matching as a Quadratic Form}
% \textcolor{teal}{[ZS: explaining $P_t\Gamma$ can be quite wordy and I'm not very content with how it is explained in the current version. We can assume $\mathbb{E}_{p_{\textrm{data}}}[\phi_k]$ is zero outside of a finite number of $\phi_n$s, but it would just be an device to simplify the narrative.]} \mr{Shall we say that it is possible to prove the equation and defer to the appendix?} 
% \textcolor{teal}{[ZS: proving the result in both Hermite and trig cases are both quite trivial. It is mainly that explaining both in a unified way seems not very ideal. I think unifying it is ultimately useful because BM semigroup is heavily featured in all illustrations]}\mr{I have attached the page from those notes you wrote at the turing, which had similar notation to here - we would need to write down (**) for this, right? }

% \begin{figure}
%     \centering
% \includegraphics[height=0.15\textheight,]{figs/turing_note.pdf}    
% \end{figure}

To formulate \ac{oism}, we make one further assumption on the spectral properties of the infinitesimal generator $\mathscr{L}$: the eigenfunctions $\{\phi_n\}_{n\in\mathcal{I}}$ form a countable Hilbertian basis in $L^2(\pi)$ \citep{bakryAnalysisGeometryMarkov2013}. 
This is the case for the \ac{ou} forward process, so that we can write $\mathcal{I} = \mathbb{N}$, while the spectrum of the \ac{bm} forward process becomes countable if its domain is modified from $\mathbb{R}^d$ to a finite interval; see \Cref{app: truncated bm} for a detailed explanation.
%\mr{perhaps mention also that this is not variance exploding anymore} \mr{Given that we anyway truncate the basis, is it really necessary that the spectrum is countable?}
%\textcolor{teal}{[ZS: yes -- the truncation is also a consequence of the finite interval property. $\exp(\imath\mu x)$ still satisfies the spectral property but it is not in the domain of $\mathscr{L}$ because its mean is not zero.}

Following the same notation from \Cref{subsec: rederive score matching}, and with $[n] = \{1 , \dots , n\}$, we aim to learn an energy-based model $f_\mbalpha(\mbx)=\sum_{k\in [n]} \alpha_k \phi_k(\mbx)$, parametrized by time-varying coefficients $\mbalpha \in \mathbb{R}^{n}$, 
% \mr{change name from $\alpha$ to $\gamma$, to avoid confusion with the coefficients of OU?} 
such that $\tilde{\mbs}_\mbalpha(\mbx)=\nabla f_\mbalpha(\mbx)$ minimizes the score matching loss \eqref{eq: score matching p0}.
%with respect to $\nabla\log\frac{\rho_t}{\pi}(\mbx)$. 
Given the eigendecomposition of $\mathscr{L}$, it is straightforward that the second term 
%\textcolor{red}{\sout{$\mathbb{E}_{p_\text{data}} \left[ P_t\mathscr{L} f_\mbalpha\right]$} 
in \eqref{eq: score matching p0} can be written in closed form, provided we know the expected values $\mathbb{E}_{\rho_0}[\phi_k]$:
\begin{align*}
    \mathbb{E}_{\rho_0}\left[P_t\mathscr{L} f_\mbalpha\right] = \sum_{k\in [n]} \alpha_k \lambda_k e^{\lambda_k t} \mathbb{E}_{\rho_0}[\phi_k] = \mbb_t^\top \mbalpha, \qquad \mbb_t^{(k)} := \lambda_k e^{\lambda_k t} \mathbb{E}_{\rho_0}[\phi_k]. 
\end{align*}
Likewise, the first term in \eqref{eq: score matching p0} is quadratic in $\mbalpha$ :
\begin{align*}
    \mathbb{E}_{\rho_0}\left[P_t \Gamma\left( f_\mbalpha\right) \right] &= \sum_{k, \ell \in [n]} \alpha_k \alpha_\ell \mathbb{E}_{\rho_0} \left[P_t \Gamma(\phi_k, \phi_\ell)\right]=\mbalpha^\top\mbA_t \mbalpha , \qquad \mbA_t^{(k, \ell)} := \mathbb{E}_{\rho_0}\left[P_t \Gamma(\phi_k, \phi_\ell)\right]. 
\end{align*}
 In general, numerical methods would be required to calculate the terms $P_t \Gamma(\phi_k, \phi_\ell)$, for example, using simulation from the forward process.  
 However, this is not the case for the \ac{ou} process, whose eigenfunctions are Hermite polynomials, or the \ac{bm} process with its associated trigonometric eigenfunctions, as in each case the product of two eigenfunctions is also an eigenfunction, albeit of a possibly higher index $m$.
 %but still finite subspace $\{\phi_h\}_{h\in[m]}$. 
%that enjoy recurrence relations, \textcolor{red}{allowing to express their products} as a linear combination of Hermite polynomials in a larger set $\left(\phi_h\right)_{h\in[m]}$, \textcolor{red}{$m>n$}. 
%\textcolor{red}{\sout{That is,} 
This feature allows us to cast $\phi_k\phi_\ell=\sum_{h\in[m]} \beta_h^{(k, \ell)} \phi_h$ as a linear combination with appropriate coefficients $\beta_h^{(k, \ell)}$, and to use the carr\'{e}-du-champ operator and \eqref{eq: eigenrelation} to obtain the explicit formula 
\begin{align*}
    \mbA_t^{(k, \ell)} &= \sum_{h\in[m]} ({ \textstyle \frac{\lambda_h-\lambda_k-\lambda_{\ell}}{2} }) e^{\lambda_h t}\beta_h^{(k, \ell)} \mathbb{E}_{\rho_0}\left[\phi_h\right].
\end{align*}
%\textcolor{red}{\sout{Therefore,}As a result,} 
Consequently, provided we know the expectations $\mathbb{E}_{\rho_0}[\phi_h]$, %\mr{do we mention how to obtain/estimate these here?} 
we can construct the score matching loss~\eqref{eq: score matching p0} as a quadratic form% \textcolor{red}{\sout{, which allows for a simple closed-formed minimizer $\widehat{\mbalpha}:=\mbA_t^{-1} \mbb_t$:}}
\begin{align}
    \mathcal{L}\left(f_\mbalpha\right) & \stackrel{+C}{=} \mbalpha^\top \mbA_t\mbalpha + 2\mbb_t^\top \mbalpha, \label{eq: oism main}
\end{align}
which is explicitly minimised at $\widehat{\mbalpha}:=\mbA_t^{-1}\mbb_t$.
A score matching estimator 
\begin{align}
\tilde{\mbs}_t(\mbx) = \sum_{k \in [n]} \hat{\alpha}_k \nabla \phi_k(\mbx) \label{eq: oism final}
\end{align} 
can therefore be learned
%with $f_t(\mbx) = -\frac{1}{2} \mbx^\top \mbA_t \mbx + \mbb_t^\top \mbx$ \textcolor{red}{Use $f_t(\mbx) = -\frac{1}{2} \mbx^\top \mbU_t \mbx + \mbv_t^\top \mbx$ instead, to avoid confusion with $\mbU_t$ and $\mbv_t$ defined above. Fix below, accordingly, and connect $\mbU_t$ and $\mbv_t$ with the time-varying $\alpha$s defined above} for all $t > 0$, 
\emph{without requiring the forward process to be simulated}, by approximating the expectations $\mathbb{E}_{\rho_0}\left[\phi_h\right]$ using averages taken over the dataset and solving \eqref{eq: oism main} to obtain the time-varying  coefficients $\hat{\mbalpha}$ appearing in \eqref{eq: oism final}.

\begin{example}[OU process, continued]
    To build intuition for how the score function is approximated in \ac{oism},  recall from \Cref{ex: ou again} that polynomials are eigenfunctions of the \ac{ou} infinitesimal generator~$\mathscr{L}$. 
For instance, one can show that, with arbitrary vector $\mbv$ and matrix $\mbU$, 
% the first order polynomial 
$\mbv^\top\mbx$ has eigenvalue $1$ and 
% the second order polynomial 
$\mbx^\top\mbU\mbx - \text{tr}(\mbU)$  has eigenvalue $2$. 
Construction of a score matching estimator \eqref{eq: oism final} based on the polynomial eigenfunctions of first and second orders, therefore, only requires computing the empirical mean $\widehat{\mbmu}$ and covariance $\widehat{\mbSigma}$ of the dataset. 
In fact, the corresponding energy-based model coincides with a Gaussian approximation of the dataset \citep{wangunreasonable}.
\end{example}
\subsection{Regularised Estimation of Eigenfunctions via Shrinkage}
\label{subsec: shrink}

%\textcolor{teal}{[ZS: this section sounds like a discussion, so there should be more pointed way to write this. I would prefer to cite Botev but it's more tangential (I think it gives the clearer perspective on why regularity is needed, but we can expand on the Stein effect of the orthogonal series to achieve the same effect)]}
Our discussion so far has shown how to convert expectations $\mathbb{E}_{\rho_0}\left[\phi_k\right]$ into explicit approximations of the score function $\tilde{\mbs}_t(\mbx)$, yet these expectations with respect to the data-generating distribution must still be estimated. 
The famous result of Stein showed that the sample mean is an inadmissible estimator for estimating three or more expectations from a dataset \citep{steinInadmissibilityUsualEstimator1956}; see \Cref{app: oism addition} for a detailed explanation.
The tensorized structure of the spectrum in high dimensions (c.f. \Cref{ex: bm spectrum} and \Cref{ex: ou again}) entails estimating a potentially large number of expectations in \ac{oism}, motivating the consideration of \emph{shrinkage} estimators, as will now be discussed.

\subsection{An Illustrative Experiment}
\label{sec: ill exp}

\begin{figure}[t!]
    \centering
    \includegraphics[width=1.0\textwidth]{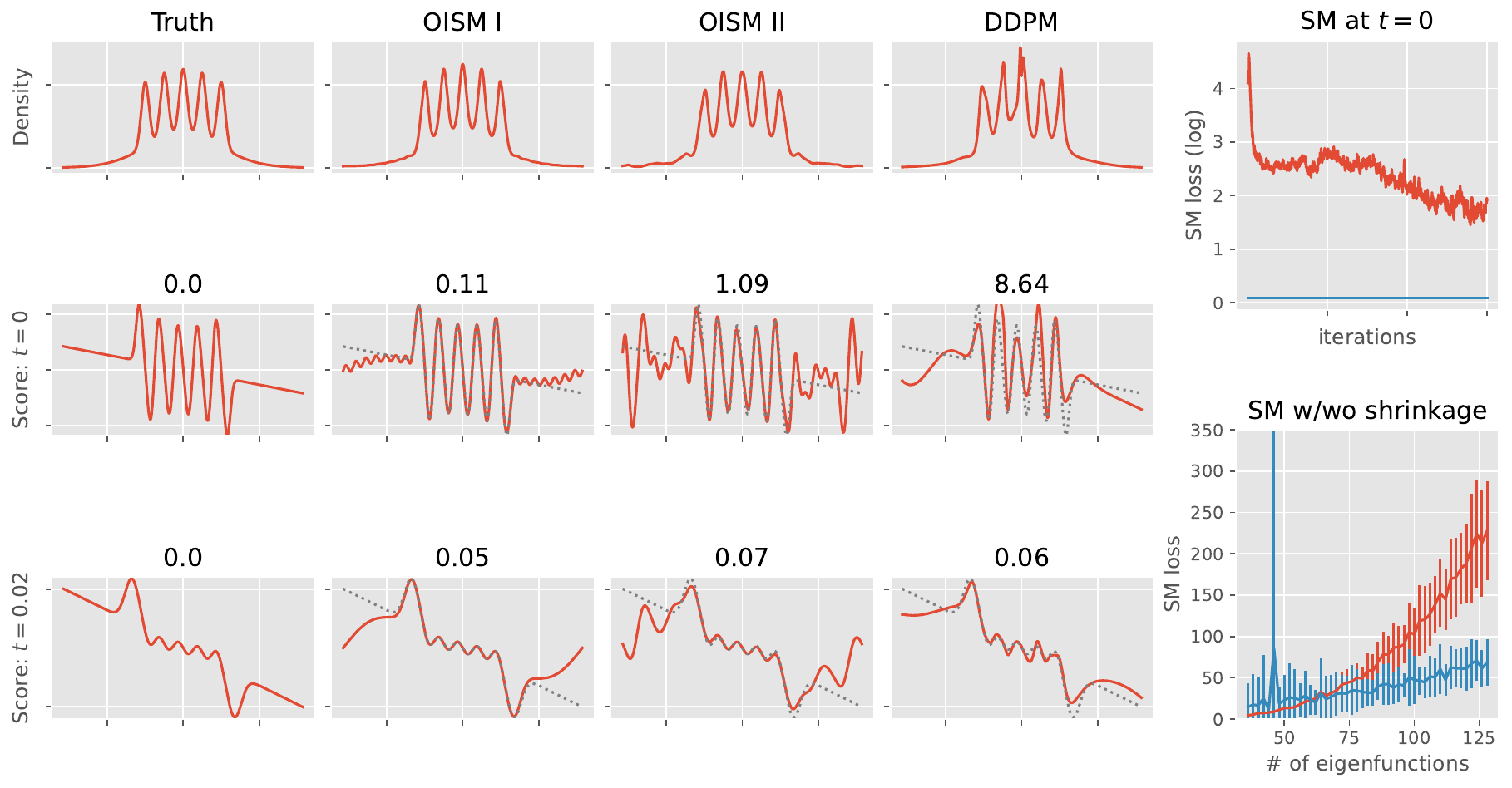}
    %\caption{An empirical proof-of-concept. 
    %The task is to construct a generative model for $p_{\text{data}}$ based on the training dataset consisting of independent samples from $p_{\text{data}}$ in \textbf{(a)}. 
    %Standard diffusing modelling (DM) produced the density estimate in \textbf{(b)}, and the samples in \textbf{(c)} were generated from this DM using the \ac{sde}-based stochastic sampler of \citet{songImprovedTechniquesTraining2020}.
    %Our proposed \emph{operator-informed score matching} (OISM) method is presented respectively \textbf{(e)}. 
    %Computational costs were equal for \textbf{(b)} and \textbf{(e)}.
    %\textcolor{blue}{[Zheyang to update this figure and caption]}
    %}
    \caption{
    Illustration of \ac{oism} in 1D. 
    The first column contains the ground truth density and score functions $\nabla \log \rho_t$ at $t=0$ and $t=0.02$. 
    \textbf{OISM I} is based on exact $\mathbb{E}_{\rho_0}[\phi_k]$, highlighting the capacity for an eigenfuntion basis to provide high-quality score estimates across different levels of noise perturbation, which leads to accurate density estimates via the probability flow \ac{ode}. 
    \textbf{OISM II} instead uses data-based shrinkage estimates for $\mathbb{E}_{\rho_0}[\phi_k]$; the density remains accurately estimated. 
    As a baseline, the fourth column showcases the score and density estimates obtained via \textbf{DDPM}, a neural network trained via denoising score matching (SM).
    [In the second and third rows the score matching loss in displayed in the subtitles.]
    The final column (top) shows the evolution of the SM loss for \textcolor{ggplot1}{\textbf{DDPM} (red)} and \textcolor{ggplot2}{\textbf{OISM II} (blue)}; and (bottom) the SM loss for \textbf{OISM II} as a function of the number of eigenfunctions, corresponding to \textcolor{ggplot1}{sample means (red)} and \textcolor{ggplot2}{shrinkage estimators (blue)}. 
    [Error bars denote the standard error over 50 simulations.]
    The Stein effect is clearly visible, as we observe little difference between sample mean and shrinkage estimators when the dimensionality of eigenfunction is low, however, the performance gap becomes clearly visible as the number of eigenfunction increases. 
    %\textcolor{green}{[CJO:  Typo in top right panel - should be ``at $t=0$'']}
    %\textcolor{teal}{[will fix shortly]}
    }
    \label{fig: toy main}
\end{figure}

 %The performance of \ac{oism} as a toy example is illustrated in \Cref{fig: toy main}. %We first validate our findings by showcasing that for low-dimensional target distributions, \ac{oism} alone is sufficient for sample generation via the reverse process. 
As a first example to illustrate \ac{oism}, considers the 1-dimensional distribution for 
% $p_{\text{data}}$ 
$\rho_0$ 
showcased in the top left panel of \Cref{fig: toy main}, called the ``Bart Simpson'' distribution \citep[Example 6.1 of ][]{wassermanAllNonparametricStatistics2005}.
This distribution is somewhat non-trivial yet it allows for scores, densities and eigenfunction expectations to be computed. 
As a dataset we draw 2,000 samples, and consider the \ac{bm} forward process with its associated eigenfunctions $\phi_k(x) = \exp(\imath k x)$ for $k \in \{1, 2, \hdots, 25\}$. 
Specifically, we opt for the \emph{modulation estimator} \citep{beranModulationEstimatorsConfidence1998}, a multivariate extension of the James-Stein estimator that 
shrinks each sample mean.
% assigns each sample mean its own linear shrinkage parameter.
%For shrinkage, we use the \emph{modulation estimator}, a multivariate extension of the James--Stein estimator, as explained in Section 8.4 of \citet{wassermanAllNonparametricStatistics2005}.
%\textcolor{blue}{[CJO:  It would be great to define it here, if there is room.]}

Results are shown in \Cref{fig: toy main}.
Details for all low-dimensional experiments are contained in \Cref{app: experiment}.
Most notably, \ac{oism} exhibits clear capacity to approximate score functions $\nabla \log \rho_t(\mbx)$ in high-density regions of $\rho_t(\mbx)$.
As the area of high density region grows with~$t$, the \ac{oism} score estimate smoothens and becomes more accurate in a wider domain. 
%This greatly alleviates the drawbacks of its less desirable behavior in low-density areas, and helps produce an accurate density estimate. 
Furthermore, we witness a substantial reduction in the score matching loss when shrinkage is applied, thus enforcing its practical value in obtaining \ac{oism}.

\begin{figure}
    \centering
\includegraphics[width=0.7\linewidth]{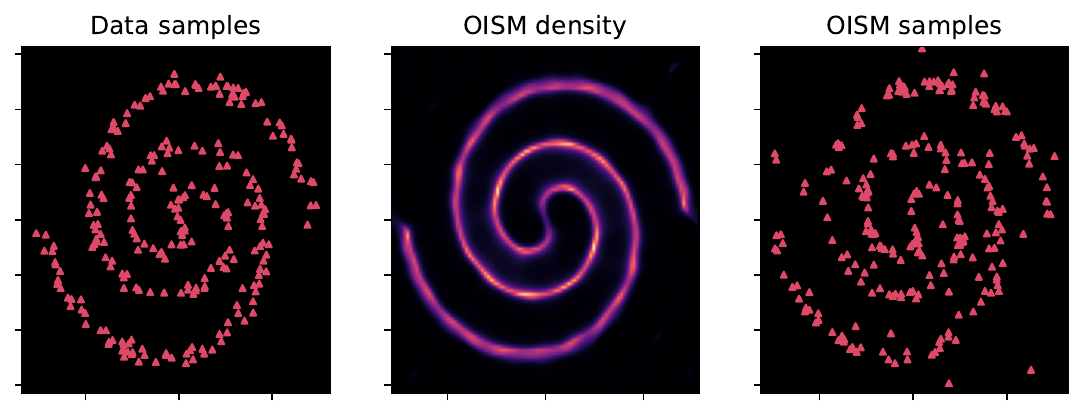}
    \caption{Illustration of \ac{oism} in 2D. The leftmost plot showcases a subset of training data. 
    \ac{oism} was applied to estimate the score function, using 200 eigenfunctions in total. 
    The density recovered by inverting the probability flow \ac{ode} \eqref{eq: prob flow ode} is quite
    accurate,
    % accurately, with only small 
    % artifacts compared to the 
    % DDPM approach of \citet{hoDenoisingDiffusionProbabilistic2020} 
    which we 
    attribute to the periodicity of the eigenfunctions of the \ac{bm} forward process. 
    %\textcolor{green}{[CJO:  The left plot doesn't look like it contains the advertised 10,000 data points, and the number 200 seems to contradict the number 125 in the main text?]}
    }
    %\textcolor{teal}{[ZS: we can't visualize 10000 data points for presentation reasons. There are 200 eigenfunctions under eigenvalue 125. ]}
    % \mr{the NN does again not so great here - was the learning rate adequate?}
    \label{fig: toy 2d}
\end{figure}

A second example, to illustrate \ac{oism} in 2D, is provided \Cref{fig: toy 2d}.
Here all trigonometric eigenfunctions with eigenvalue $\lambda_n\leq 125$ are enumerated, yielding a score estimate consisting of 200 eigenfunctions in total, and a dataset of size 10,000 was used. 
%A relatively large number (10k) of samples are required in order to produce reliable expectation. 
For this low-dimensional target, the `pure' \ac{oism} methodology together with shrinkage is sufficient to learn an accurate generative model.
This is not the case in high-dimensions, where shrinkage estimation becomes more complicated and we instead advocate for \ac{oism} as a useful addition to neural network-based methods; see \Cref{subsec: practical}. 

\subsection{Practical \ac{oism} in High Dimensions}
\label{subsec: practical}

\ac{oism} can be a powerful alternative to standard \ac{dm} in the low-dimensional setting, as showcased in \Cref{sec: ill exp}.
However, a truncated eigenbasis cannot fully replace flexible learning of the score function when the data are high-dimensional, since these features (i.e. the eigenfunctions) are not adapted to the dataset.
Our proposal to extend \ac{oism} to high-dimensional settings is quite simple: we construct a first approximation $\tilde{\mbs}_t(\mbx)$ to the score function as just described, and learn the \emph{residual} $\mbr_t(\mbx) := \nabla \log \frac{\rho_t}{\pi}(\mbx) - \tilde{\mbs}_t(\mbx)$ using a neural network trained via denoising score matching \citep{hoDenoisingDiffusionProbabilistic2020}.
This set-up is appealing; it represents a minimal modification of existing code, which can be achieved by augmenting the output 
% \mr{this is vague/wrong if claim we learn the residual?} 
of the score function approximator in the \ac{dm}.

An important issue is \emph{which eigenfunctions to include} in the first approximation $\tilde{\mbs}_t(\mbx)$.
This question is rather difficult, and we propose a simple heuristic in this work while leaving the idea of data-adaptive eigenfunction selection for future work.
Namely, we note that if the distribution $\rho_0$ can be factorized, then one would only need to estimate expectations of the \emph{univariate} eigenfunctions of the form $\mathbb{E}_{\mbx\sim\rho_0}\left[\phi_k(x_i)\right]$, $i \in \{ 1, \dots, d\}$, where $x_i$ denotes the $i$th component of $\mbx$; c.f. \Cref{ex: ou again}. 
%Working under the factorized assumption, the score estimate 
% $\tilde{\mbs}^{\text{\ac{oism}}}(\mbx) := \widehat{\mbalpha}^\top \nabla\mbphi(\mbx)$ 
%\eqref{eq: oism final}
%coincides a total of $d$ 1-dimensional score functions. 
Our heuristic is to use the first $n$ univariate eigenfunctions in each data dimension \emph{without} shrinkage, denoting this \textbf{OISM ($n$)}; by ignoring interaction terms, the need for shrinkage may be reduced, and empirical support for this heuristic is provided in \Cref{sec: empirical}.

%where it is seen to outperform standard \ac{dm}. 
%Full experimental details can be found in \Cref{app: experiment}.
%In addition, we showcase in \Cref{app: mnist} that in a higher-dimensional setting, learning the residual scores alongside a linear score estimator yields similar results while providing marginal improvements on the deterministic simulation of samples using the probability flow \ac{ode} \eqref{eq: prob flow ode}. 
%The \ac{sde}-based stochastic sampler of \citet{songImprovedTechniquesTraining2020} can be seen as the sum of \eqref{eq: prob flow ode} and a Langevin diffusion converging to $\rho_t$ at each time $t$; linear \ac{oism} achieves the same effect by enforcing convergence towards a Gaussian approximation of $\rho_t$.
%Higher-order score approximations has the potential to give better-informed closed-form score estimates, and is investigated in \Cref{sec: empirical}. 

%

%% file: sections/experiments.tex
\section{Experimental Assessment}
\label{sec: empirical}

This section presents an empirical assessment of \ac{oism} on a prototypical high-dimensional generative modelling task.
For this assessment we used the CIFAR-10 dataset, which consists of 60,000 colour images of dimension $32 \times 32$, partitioned into 50,000 training and 10,000 test images \citep[][Chapter 3]{krizhevsky2009learning}.
%in 10 classes, with 6000 images per class. 
%There are 50000 training images and 10000 test images \citep[][Chapter 3]{krizhevsky2009learning}.

% \textcolor{blue}{[CJO:  We can add another dataset later]}\mr{We might not have the time to run ImageNet before next week, unless it is for low order polynomials and not so many itarations. }

\paragraph{Implementation detail}
%\textcolor{blue}{[CJO:  Details to be confirmed - explain that we are using the variance-preserving model]}
Our implementation makes use of the \ac{ou} (variance-preserving) process \eqref{eq: OU process} as the noising process for high-dimensional simulation, following the weighted noising schedule as established by \citet{songMaximumLikelihoodTraining2021}, and the neural network architecture DDPM++ proposed by \citep{songScoreBasedGenerativeModeling2020}. 
The eigenfunctions of the \ac{ou} process are Hermite polynomials, and we considered a maximal (univariate) polynomial order ranging from 2 to 6. 
We opt for the simple solution of sample mean with no consideration of shrinkage, as we did not observe an improvement when shrinkage was applied in our testing (not shown). 
This finding indeed seems consistent with the lack of Stein's effect as shown in \Cref{fig: toy main}.
The rest of our training procedure was identical to the description in \citep{songScoreBasedGenerativeModeling2020}. 
Code to reproduce these experiments can be obtained at \url{[blinded]}. 
Computation was performed on A100 GPUs and training of each model using a total of 500,000 iterations required approximately 30 hours in total. 
There was no increased cost at training time to using \ac{oism} as all linear systems involving $\mbA_t$ and $\mbb_t$ were pre-solved.
%\textcolor{teal}{[ZS: from my point of view the explanation of the quantization, number of iters can be moved to the appendix. ]}
%\textcolor{orange}{[HH:For high-dimensional simulations, we mainly focus on the OU process (\emph{variance-preserving}  \ac{dm}), so we employ \sout{For the neural network we employed} DDPM++ (cont) \citep{songScoreBasedGenerativeModeling2020} as the neural network. The same structure is used as the complementary part to the operator, but the order of the Hermite polynomials is varied from 2 to 6 to observe the impact of different orders.
%We follow the same training procedure described in \citep{songScoreBasedGenerativeModeling2020}, yet reduce the number of iterations to 500,000 for training efficiency. Furthermore, since pixelated data lacks a density ratio with respect to the Gaussian density, learning a diffusion model directly on such data constitutes an ill-posed problem. So we introduce a uniform noise to the training data as well. Unlike the 2D setting, we don't consider shrinkage here since finding a better shrinkage for coefficients computed from the high-dimensional linear system is hard. Code to reproduce these experiments can be obtained at \url{[blinded]}

%\paragraph{Training cost}
%\textcolor{orange}{[HH:\sout{For training the neural network we follow}]}  \citet{songScoreBasedGenerativeModeling2020}.
%All models are trained on A100 GPUs. 
%\textcolor{blue}{[CJO: Or, if we do something different, then here is where we explain what we did - may not need the exponential moving average stuff.]}

\begin{figure}[t!]
    \centering
    \includegraphics[width=\linewidth]{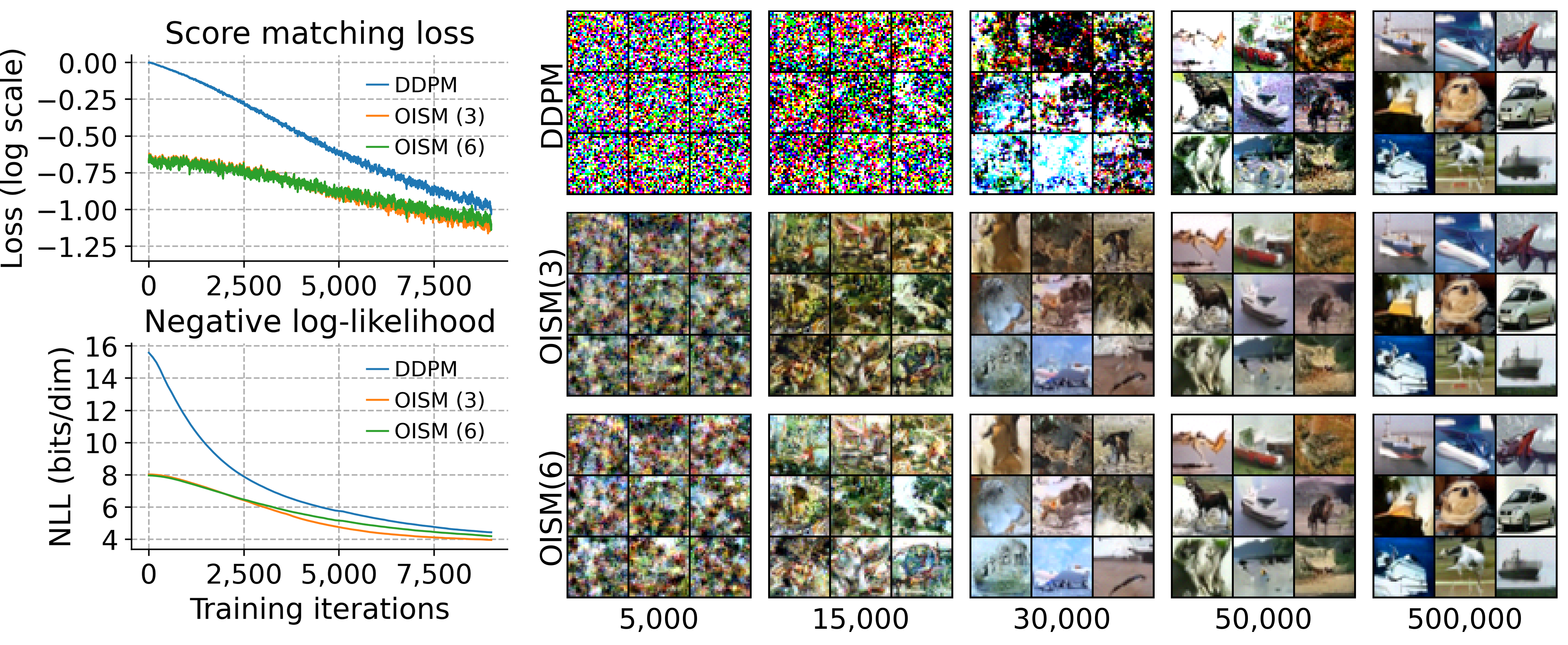}
    \caption{Performance evaluation on CIFAR-10. 
    Top left: Smoothed score matching (SM) loss for \textbf{DDPM} \citep[the maximum likelihood estimator of][]{songMaximumLikelihoodTraining2021}, \textbf{OISM} (3), and \textbf{OISM} (6). 
    Bottom left: Negative log-likelihood evaluated on the test dataset. 
    Right: Samples corresponding to different training iterations, obtained by reversing the probability flow ODE~\eqref{eq: prob flow ode}.}
    \label{fig: cifar10}
\end{figure}

\paragraph{Results}
Full results are shown for \textbf{OISM} (3) and \textbf{OISM} (6) in \Cref{fig: cifar10}, with further results and discussion reserved for \Cref{app: mnist}.
In both cases, \ac{oism} significantly accelerates training, providing an informative initial guess of the score function, and, by extension, a rough estimate of the data distribution. 
Comparing the the score matching loss, log-likelihood and generated samples displayed in \Cref{fig: cifar10}, we see that the initial several thousand iterations can effectively be skipped over entirely with the help of \ac{oism}. 
On the other hand, higher order does not guarantee better performance, with \textbf{OISM} (6) performing slightly worse than \textbf{OISM} (3).
%The \ac{oism} in low-density regions (see figure \Cref{fig: toy main} provides a counterintuitive score estimate, which would be exacerbated as the order of polynomial increases. Further exploration can be found in the supplement. 
%Our experiments confirm that the operator does inform some information compared with a random Gaussian noise, which serves as a better initial guess.

%\textcolor{teal}{[ZS: I don't like "huge linear system" as an explanation -- we don't know for sure if the linear solver was working properly and it is out of our control but we could address this issue.]}

%\textcolor{orange}{[HH: From the evaluation loss and NLL, incorporating an operator eases the training process much since we start from a lower training loss or NLL. In other words, the first several thousand iterations can be avoided entirely, which therefore accelerates the training. The generated samples support our findings further, where we could see clearer samples earlier than DDPM. However, higher order does not mean better metrics performance since eigenfunctions scale with the dimension of the data and the maximum order of the polynomial, which results in a huge linear system. However, with iterations, we could still find better samples compared to the lower-order polynomials. Further exploration can be found in the supplement. Our experiments confirm that the operator does inform some information compared with a random Gaussian noise, which serves as a better initial guess.]}
\paragraph{Interpretation}
Even a low-order approximation, such as \textbf{OISM} (3), has the potential to substantially accelerate training the generative \ac{dm}.
However, and somewhat counter-intuitively, a benefit from using higher-order approximations was not observed.
%To understand this, we note that the results in \Cref{fig: cifar10} were obtained without the shrinkage estimation discussed in the low-dimensional context in \Cref{subsec: shrink}. 
%Indeed, in testing (not shown) we did not observe an improvement when shrinkage was applied.
This counterintuitive result might stem from a lack of understanding of the interplay between our novel score estimates and the implicit regularisation provided by the neural network. 
One can speculate that \textbf{DDPM} outperforms \ac{oism} in low-density regions thanks to its inductive bias, as observed in \Cref{fig: toy main}. 
Indeed, as the order of polynomial increases in \ac{oism}, the neural network might be forced to fit a residual that has greater complexity and curvature in the tail, for which the usual linear extrapolation of the neural network may be ill-suited.

%% file: sections/post_discussion.tex
\section{Discussion}
\label{sec: discuss}

This paper contributes to an improved theoretical understanding of \acp{dm}, demonstrating the potential of 
%connections to diffusion operators, kernel embeddings, and density ratio estimation; as a by-product, the novel \ac{rdks} and 
\ac{oism} as a practical tool to accelerate training of Markov \acp{dm}.
In this section we discuss related work, and the limitations and potential of the methods we have proposed.

\paragraph{Related work}
Outside the scope of generative modeling, our work closely relates to the use of orthogonal series in density estimation. \citet{whittleSmoothingProbabilityDensity1958,cencov1962estimation,schwartzEstimationProbabilityDensity1967} observed a density estimate could be obtained provided that $\mathbb{E}_{\rho_0}[\phi_k]$ were known, and its error can be improved with the introduction of shrinkage estimators; see \citet{efromovichOrthogonalSeriesDensity2010}. Within diffusion models, %Two recent papers have discussed the interplay between the forward process and score matching in detail. 
\citet{scarvelisClosedFormDiffusionModels2023} considered a simple forward process taking the form of a weighted mixture of the data distribution $p_{\text{data}}$ and the noise distribution $\pi$, so that both the forward process and the associated scores can be exactly computed.
This approach circumvents neural approximation of score functions, but it memorizes the finite dataset when applied naively. Our regularized feature-based approach obviates the need for iterating the dataset and alleviates the issue of memorization. 
The loss \eqref{eq: score matching p0} 
% also appeared in 
relates to the loss proposed by 
\cite{bentonDenoisingDiffusionsDenoising2024}, where Markov operators are used  
as a means to generalise Markov \acp{dm} to Riemannian manifolds, but the authors of that work did not consider \eqref{eq: score matching p0} as a tool to derive explicit approximation to the score function, as instead proposed in this work.
%introduces a performance gap relative to the state-of-the-art for tasks such as image generation.

%Our contribution differed in that we focused on standard Markov \acp{dm} and investigated in detail theoretical aspects of memorization and smoothing, developing a systematic methodology for smoothing in the \ac{dm} context. 
%
%
%
%Second, \citet{laiImprovingScorebasedDiffusion2023} considered a Markov \ac{dm} and observed that the score functions associated to the noise-perturbed distributions satisfy a \ac{fpe}.
%The authors then developed a regularized version of score-matching in which the score \ac{fpe} is approximately enforced; we showed in \Cref{app: consequences} that score \ac{fpe} can be expressed in terms of Markov diffusion operators, providing additional insight into this method.  
%Our approach differed to this work in that we further leveraged the spectral properties of Markov diffusion operators to obtain novel score estimators that \emph{automatically} conform to the constraints of a Markov \ac{dm}.

%\textcolor{red}{Our derived loss can be related to the DSM loss of \cite{bentonDenoisingDiffusionsDenoising2024}, but they don't use it derive a new score estimate, but to work on Riemmanian manifolds.} 
%
%

\paragraph{Limitations and potential}
This work presented a novel viewpoint of encoding diffusion operators into score-based generative modeling; however, our investigation was mostly theoretical and our empirical assessment was based on `vanilla' \acp{dm} rather than the state-of-the-art.
This was a deliberate decision, to emphasise only the relevant aspects of \acp{dm}; further investigation will be required to explore the performance of \ac{oism} for more sophisticated \acp{dm} on more challenging generative modelling tasks. 
Though low-order approximations like \textbf{OISM} (3) appear to be powerful and straightforward to implement, an interesting question is whether further performance gains can be achieved using higher-order approximations; we speculate this is possible via appropriate shrinkage estimation or data-adaptive choice of eigenfunctions.
Indeed, increasing shrinkage applied to the coefficients of higher-order eigenfunctions in \textbf{OISM} (6) would increase performance at least to the level of \textbf{OISM} (3), so the development of effective shrinkage strategies is a promising direction for future work.
%The proposed \ac{rdks} method can in principle circumvent neural score approximation in Markov \acp{dm}, but relies on estimation of the metric tensor, i.e. the data manifold.
%Further work will be required to investigate whether Riemannian \ac{dks} can be made to work in a high-dimensional context.
%The linear version of \ac{oism} is immediately applicable and demonstrated practical potential, but further work (e.g. tensorization of Hermite polynomials) will be needed to make higher-order approximation of the score function practical. 
Despite these limitations, our contribution supports the view that Markov diffusion operators provide valuable insight into \ac{dm}, and have the potential both to improve existing methods and inspire new methodological development.

%% file: sections/ack.tex
ZS and CJO were supported by EP/W019590/1.
CJO was supported by a Philip Leverhulme Prize PLP-2023-004.
Computation was performed using \citet{CREATE2025} at King's College London, UK.

%% file: sections/supplement_ve.tex
\section*{\Large Appendices}

These appendices supplement the paper \textit{Operator-informed score matching for Markov diffusion models}.

\Cref{app: brownian} presents the Markov diffusion operators for a more general class of \acf{lti} \acp{sde}.
\Cref{app: truncated bm} takes a closer look at the peculiarities involving \ac{bm}, and proposes a truncated variant to address them. 
\Cref{app: oism addition} revisits the use of orthogonal functions in nonparametric density estimation, and lays out the details about the specific shrinkage estimator, as well as the use of preconditioner for numerical stability. 
\Cref{app: consequences} explores further consequences of Markov diffusion operators in the context of related work, including time score matching \citep{choiDensityRatioEstimation2022} and the score \ac{fpe} \citep{laiImprovingScorebasedDiffusion2023} and the generalized score matching \citep{bentonDenoisingDiffusionsDenoising2024}.
%Related work on the use of orthogonal series for density estimation is reviewed in \Cref{app: series density}.
Full details required to reproduce our two-dimensional experiment are contained in \Cref{app: experiment}, and full details to reproduce our image generation experiments are contained in \Cref{app: mnist}.

\section{Diffusion Operators for Markov Forward Processes}
\label{app: brownian}

This appendix presents Markov diffusion operators for a wider class of \acf{lti} \acp{sde}, generalizing beyond the \ac{ou} and \ac{bm} processes featured in the main text.
Chapter 6 of \citet{sarkkaAppliedStochasticDifferential2019} contains helpful explanations on this topic. \ac{lti} \acp{sde} take the form of the following It\^{o} diffusion:
\begin{align}
    \d \mbX_t = \mbF \mbX_t + \mbL \d\mbW_t,  \label{eq: lti}
\end{align}
for which existence and uniqueness of the solution is guaranteed for any choice of the drift matrix~$\mbF$ and of the diffusion matrix $\mbL$, while its stability, and existence of and convergence to a stationary distribution $\pi$ depend on these matrices.
The above formula covers most diffusion forward processes, for example \ac{bm} is obtained with $\mbF=\mathbf{0}, \mbL=\sqrt{2}\mbI$, \ac{ou} with $\mbF=-\mbI, \mbL = \sqrt{2}\mbI$, and the critically-dampened OU process
 with 
$$
\mbF=\left(\begin{matrix}\mathbf{0} & \mbI\\ -\mbI & -c\mbI\end{matrix}\right), \qquad \mbL=\left(\begin{matrix}\mathbf{0} & \mathbf{0}\\ \mathbf{0} & \sqrt{2c}\mbI\end{matrix}\right), 
$$
corresponding to recent efforts of dampening the forward processes with momentum variables, see \citet{dockhornScoreBasedGenerativeModeling2021}.
The stationary measure and Markov semigroup associated with \eqref{eq: lti} are discussed in \Cref{subsec: stationary measure}.
The infinitesimal generator and carr\'{e}-du-champ operators are discussed in \Cref{subsec: gen and cdc}, and the spectrum of the infinitesimal generator for a wide class encompassing the \ac{ou} process is discussed in \Cref{app: spectrum app}.
%The case of a Riemannian Langevin diffusion is discussed in \Cref{app: pointers}.
Lastly, some useful properties of Hermite polynomials (the eigenfunctions of the infinitesimal generator of the \ac{ou} process) are contained in \Cref{app: hermite}.

\subsection{Characterizing the Stationary Measure}
\label{subsec: stationary measure}

The existence of a finite stationary measure $\pi$ for \eqref{eq: lti} is determined by the eigenvalues of $\mbF$. If the real parts of the eigenvalues of $\mbF$ are strictly negative, then \eqref{eq: lti} evolves towards $\pi$ a zero-mean Gaussian distribution $\cN(\mathbf{0}, \mbSigma_\infty)$, 
where the covariance matrix $\mbSigma_\infty$ satisfies the  Lyapunov equation (eq. 6.69 of \citet{sarkkaAppliedStochasticDifferential2019}):
\begin{align*}
    \mbF\mbSigma_\infty + \mbSigma_\infty\mbF^\top + \mbL\mbL^\top = \mathbf{0}.
\end{align*}
Furthermore, the conditional distributions of \eqref{eq: lti} have the following closed form solution: 
\begin{align}
    \rho_t\left(\tilde{\mbx}\vert\mbx\right) &= \mathcal{N}\left(\tilde{\mbx}\vert\mbS_t\mbx, \mbSigma_t\right), \quad \mbS_t := \exp(t\mbF), \quad \mbSigma_t := 
\mbSigma_\infty-\mbS_t\mbSigma_\infty\mbS_t^\top. \label{eq: lti marginal}
\end{align}
While a thorough analysis of the spectral decomposition of \ac{lti} \acp{sde} can be complicated, its Markov transition kernel, as described in \eqref{eq: lti marginal}, remains conditionally Gaussian. 
%gives a closed from \ac{oism} solution, equivalent to approximating the data distribution $\rho_0$ with a Gaussian \citep{wangunreasonable}. %\mr{This last sentence needs grammar fix, but it is also too implicit, could you explain further?}
%\mr{State that these do not hold for the \ac{bm}, that has exploding variance. Or do they formally apply?}

\subsection{Infinitesimal Generators and Carr\'{e}-du-Champ Operators}
\label{subsec: gen and cdc}
%\textcolor{teal}{[ZS: I will rewrite part of the following sections. LTIs are a bit complicated when the associated Markov semigroup is not symmetric, but \citet{pavliotisStochasticProcessesApplications2014} is a useful reference]}\mr{Ok, but doesn't it hold that the semigroup is symmetric in $L^2(\pi)$? We could stress here that we use this in our computation, and therefore consider as if it was symmetric}
%\textcolor{teal}{[ZS: I think it is nice to stress that LTIs can be analyzed theoretically using Hermite polynomials, which leads to extensions to asymmetric semigroups, but the integration by parts becomes more complex. ]}
Denoting $\mbG:=\frac{1}{2}\mbL\mbL^\top$, the infinitesimal generator takes the following form:
\begin{align}
\mathscr{L} f (\mbx) = \langle \mbF\mbx, \nabla f(\mbx)\rangle + \langle \nabla, \mbG \nabla f(\mbx)\rangle. \label{eq: lti generator}
\end{align}
While the generator \eqref{eq: lti generator} applies to all \ac{lti} \acp{sde}, our methodology is most applicable when the associated Markov semigroup  is \emph{symmetric}. We follow Definition 1.6.1 of \citet{bakryAnalysisGeometryMarkov2013}: 
\begin{definition}[Symmetric Markov semigroup]
    A Markov semigroup $(P_t)_{t\geq 0}$ is \emph{symmetric} with respect to the invariant measure $\pi$, if for all $f, g \in L^2(\pi)$ and $t\geq 0$
    \begin{align*}
        \int f(\mbx) P_t g(\mbx)\mathrm{d}\pi(\mbx) = \int g(\mbx) P_t f(\mbx)\mathrm{d}\pi(\mbx).
    \end{align*}
\end{definition}
The Markov semigroup of an \ac{lti} is symmetric when $-\mbF$ and $\mbL$ are positive definite matrices, which excludes %irreversible \mr{irreversibility not defined before} 
forward processes with a more complex structure, such as underdamped systems \citep{dockhornScoreBasedGenerativeModeling2021}. We could infer symmetry by checking if $k_t(\mbx, \mby):=\frac{\rho_t(\mby\vert\mbx)}{\pi(\mbx)}$ is symmetric. For instance, the \ac{ou} forward process has a Markov transition kernel $\rho_t(\mby \vert \mbx) = \cN(\mby\vert e^{-t}\mbx, (1-e^{-2t})\mbI)$, which gives the transition kernel w.r.t. $\pi$ as the \emph{Mehler kernel}
\begin{align*}
    k_t(\mbx, \mby) = \frac{1}{(1-e^{-2t})^{d/2}}\exp\left(-\frac{\norm{e^{-t} \mbx}^2+\norm{e^{-t} \mby}^2-2 e^{-t}\langle \mbx, \mby\rangle}{2(1-e^{-2t})}\right).
\end{align*}
When expressing $P_t$ as a kernel integral operator, the symmetry becomes evident
\begin{align*}
    \int f(\mbx) P_t g(\mbx) \mathrm{d}\pi(\mbx) = \iint f(\mbx) k_t(\mbx, \mby) g(\mby) \mathrm{d}\pi(\mby)\mathrm{d}\pi(\mbx) = \int g(\mby) P_t f(\mby) \mathrm{d}\pi(\mby).
\end{align*}
%\mr{this seems abrupt here, perhaps bring back the connection between kernel and $P_t$ from the first version of the paper, but then make symmetry a separate subsection}
%Here we characterise the infinitesimal generators and carr\'{e}-du-champ operators for \acp{sde} of the form \eqref{eq: lti}, based on relevant results from Sections 1.10 and 1.11 of \citet{bakryAnalysisGeometryMarkov2013}. 
Once we have established the symmetry of $P_t$, we could apply the same integration-by-parts formula used in the paper. 
%\mr{I now understand better why you wanted the vanishing condition before. It could be helpful to derive how symmetry implies generalized integration by part, if there is time} 
%\mr{Similarly, what might be missing, is the derivation of the intermediate steps in equations 8 - 10, in the case of symmetry, ie how do we practically use it in the end? }
The carr\'e-du-champ operator of \ac{lti} \acp{sde} can be expressed as:
\begin{align*}
    \Gamma(f, g) = \nabla f^\top \mbG \nabla g = \frac{1}{2} \big\langle \mbL^\top\nabla f, \mbL^\top\nabla g\big\rangle. 
\end{align*}
%\textcolor{teal}{[ZS: the part about BM semigroup is not that interesting, and we have mentioned in the main text, so I'm in favor of just deleting it. ]}\mr{ok}
%It is worth noting that the above expressions of the two operators apply also to the \ac{bm} despite it does not have a \textcolor{red}{\sout{``stationary measure''} finite stationary measure}. \textcolor{red}{\sout{In fact} Furthermore}, \mr{I am not sure I follow the connection here, if it is supposed to read `In fact'} the Lebesgue integration by parts can be illustrated using the Dirichlet operator and the self-adjointness \mr{not  defined. Either stick to symmetry, or define the adjoint, if making a new sub-section devoted to symmetry} of the \ac{bm}, as for all functions $f, g$ such that $\lim_{\norm{\mbx}\rightarrow\infty} f(\mbx)=\lim_{\norm{\mbx}\rightarrow\infty} g(\mbx)=0$:
%\begin{align*}
%    \cE(f, g) = \int \langle \nabla f(\mbx), \nabla g(\mbx)\rangle \d\mbx = -\int f(\mbx)\mathscr{L} g(\mbx)\d\pi(\mbx) = -\int f(\mbx)\Delta g(\mbx)\d\mbx.
%\end{align*}
%
%
\subsection{The Spectrum of 
% $\mathscr{L}$
the Infinitesimal Generator
}
\label{app: spectrum app}
When \eqref{eq: lti} corresponds to a generalized version of the Ornstein-Uhlenbeck process, that is, $\mbF=-\theta\mbI$, $\mbL=\sigma\mbI$, for some $\theta, \sigma > 0$, the stationary distribution $\pi$ becomes $\cN(\mathbf{0}, \frac{\sigma^2}{2\theta}\mbI)$. In one-dimensional cases, the spectrum of its generator is formed by the eigenpairs 
\begin{align*}
    \left\lbrace \theta n, \frac{1}{\sqrt{n!}}\mathrm{He}_n\left(\sqrt{\frac{\theta}{2}}\sigma x\right)\right\rbrace_{n \geq 0}; 
\end{align*}
see Theorem 4.2 of \citet{pavliotisStochasticProcessesApplications2014}. In high dimensions, the eigenfunctions are the tensorized form of input-transformed Hermite polynomials 
which form a basis in $L^2(\pi)$. %\mr{check if this new version contrasting 1 and multi dim reads ok. It might be helpful to give a reference or derivation ia semigroup, like done for the BM, but only if there is time.}

\subsection{Asymmetric Markov semigroups}
\label{app: asymmetry}
%\textcolor{teal}{[ZS: I did not find the exact form $\mathscr{L}^*$ in any textbook (most textbooks characterize $\mathscr{L}^*$ with respect to the Lebesgue measure), but it looks quite useful.]}
%\textcolor{teal}{[ZS: I want to draw a connection to $\frac{\rho_t}{\pi} = P_t\frac{\rho_0}{\pi}$ but I don't like the current star notation. It is quite clear that $\mathscr{L}^*$ is the Hermitian adjoint, but $P_t^*$ is not the adjoint of $P_t$. What would be a good notation do differentiate between adjoint and dual?]}\mr{In Bakry the dual is called adjoint, but then they say it has the dual interpretation. Perhaps $P_t^{\dagger}$}
When a Markov semigroup violates the symmetry constraint, the introduction of the \emph{dual} and \emph{adjoint} operators becomes necessary. Given the the infinitesimal generator $\mathscr{L}$ \eqref{eq: lti generator}, the \emph{Hermitian adjoint} of $\mathscr{L}$ in the inner product space $L^2(\pi)$ is defined as the linear operator $\mathscr{L}^*$ such that
\begin{align*}
    \forall f \in L^2(\pi),\, g \in \mathcal{D}(\mathscr{L}), \qquad \int f(\mbx)\mathscr{L}g(\mbx) \mathrm{d}\pi(\mbx) = \int g(\mbx) \mathscr{L}^* f(\mbx)\mathrm{d}\pi(\mbx).
\end{align*}
Using the integration-by-parts formula, we can determine that the adjoint generator has the following form
\begin{align}
    \mathscr{L}^* f(\mbx) := \langle \mbSigma_\infty^{-1}\mbx, \mbF \mbSigma_\infty\nabla f(\mbx)\rangle + \langle \nabla, \mbG \nabla f(\mbx)\rangle. \label{eq: asymmetric adjoint gen}
\end{align}
$\mathscr{L}^*$ is useful in defining the \emph{dual semigroup} $\left(P_t^\dag\right)_{t\geq 0}$, which acts on the set of probability measures. The  semigroup $P_t^\dag$ is implicitly defined by the duality below:
\begin{align*}
    \int P_t f(\mbx)\mathrm{d}\rho_0(\mbx) = \int f(\mbx) \mathrm{d}\rho_t(\mbx) = \int f (\mbx)\mathrm{d}\left(P_t^\dag \rho_0\right)(\mbx).
\end{align*}
Provided that $\rho_0$ is the law of $\mbX_0$, $P_t^\dag\rho_0$ is the law of $\mbX_t$. When the Markov semigroup is symmetric and the density ratio $\frac{\rho_0}{\pi}$ is within the domain of $P_t$, $P_t^\dag\rho_0$ is formally $\pi\cdot P_t\frac{\rho_0}{\pi}$, as evidenced by the following relation 
\begin{align*}
    \int f (\mbx)\mathrm{d}\left(P_t^\dag \rho_0\right)(\mbx) = \int P_t f(\mbx)\frac{\rho_0}{\pi}(\mbx)\mathrm{d}\pi(\mbx) = \int f(\mbx) \mathrm{d}\left(\pi\cdot P_t\frac{\rho_0}{\pi}\right)(\mbx).
\end{align*}
When the Markov semigroup is no longer symmetric, we can use $\left(e^{t\mathscr{L}}\right)^* = e^{t\mathscr{L}^*}$ for the derivation below.
\begin{align*}
    \int f (\mbx)\mathrm{d}\left(P_t^\dag \rho_0\right)(\mbx) =& \int P_t f(\mbx)\frac{\rho_0}{\pi}(\mbx)\mathrm{d}\pi(\mbx)\\
    =& \int e^{t\mathscr{L}} f(\mbx)\frac{\rho_0}{\pi}(\mbx)\mathrm{d}\pi(\mbx) = \int f(\mbx) \left(e^{t\mathscr{L}}\right)^*\frac{\rho_0}{\pi}(\mbx)\mathrm{d}\pi(\mbx)\\
    =& \int f(\mbx) e^{t\mathscr{L}^*}\frac{\rho_0}{\pi}(\mbx)\mathrm{d}\pi(\mbx) = \int f(\mbx)\mathrm{d}\left(\pi\cdot e^{t\mathscr{L}^*}\frac{\rho_0}{\pi}\right)(\mbx).
\end{align*}
Therefore, the dual semigroup can be similarly expressed with the adjoint operator: $P_t^\dag\rho_0(\mbx) = \pi(\mbx)\cdot e^{t\mathscr{L}^*}\frac{\rho_0}{\pi}(\mbx)$, or more concisely, 
\begin{align*}
    \frac{\rho_t}{\pi} = e^{t\mathscr{L^*}} \frac{\rho_0}{\pi}.
\end{align*}
Differentiating both sides with respect to $t$, we obtain a \ac{pde} that describes the time evolution of the density ratio $\frac{\rho_t}{\pi}$, which naturally leads to the \emph{Fokker-Planck equation}:
%\mr{different to the one for the marginal (not the ratio) or to the symmetric case? I don't think we formulated FP before this point}  form of the Fokker-Planck equation
\begin{align}
    \frac{\partial}{\partial t} \frac{\rho_t}{\pi} = \mathscr{L}^* \frac{\rho_t}{\pi}, \qquad \frac{\partial}{\partial t} \rho_t = \pi \mathscr{L}^* \frac{\rho_t}{\pi}. \label{eq: operator fpe}
\end{align}
%\textcolor{red}{where  we recover  $P_t\rho_0$ in the case $P_t$ is symmetric, which implies self-adjointness ($P_t^* = P_t$)}. On the other hand, the Hermitian adjoint $\mathscr{L}^*$ in the space of $L^2(\pi)$ takes the following form 
%\begin{align}
%    \mathscr{L}^* f(\mbx) := \langle \mbSigma_\infty^{-1}\mbx, \mbF \mbSigma_\infty\nabla f\rangle + \langle \nabla, \mbG \nabla f\rangle.
%\end{align}
%The definition of adjoint operators provides another version of integration by parts: $\int f(\mbx) \mathscr{L} g(\mbx)\mathrm{d}\pi(\mbx) = \int g (\mbx)\mathscr{L}^*f(\mbx) \mathrm{d}\pi(\mbx)$. The time evolution of the density ratio $\frac{\rho_t}{\pi}$ takes a simple expression $\frac{\partial}{\partial t} \frac{\rho_t}{\pi} = \mathscr{L}^* \frac{\rho_t}{\pi}$, analogous to the Fokker-Planck equation. \mr{I think here we can get rid of the adjoint, this is needed in the Fokker-plank of the marginals, but not for the density ratio, even if the markov semigroup is not symmetric?} Similarly, we can use the exponential operator to express the density ratio as $\frac{\rho_t}{\pi} = e^{t\mathscr{L}^*} \frac{\rho_0}{\pi}$. We illustrate the expressions above to highlight the similar expression of the density ratio as $\frac{\rho_t}{\pi}=P_t\frac{\rho_0}{\pi}$ when the Markov semigroup is symmetric, which often comes up in our derivation. 

When the Markov semigroup lacks symmetry, we cannot use the same derivation adopted in this paper that exploits the carr\'{e}-du-champ operator, and the generator $\mathscr{L}$ usually has a more complex, though not fully intractable, spectral structure. For example, the eigenfunctions still take a polynomial form, and whether or not the eigenvalues of $\mathscr{L}$ are complex-valued depends on whether the system is underdamped or overdamped; see Section 6.3 of \citet{pavliotisStochasticProcessesApplications2014} for a full explanation. 

\subsection{Hermite Polynomial Recurrences}
\label{app: hermite}
\ac{oism} for the \ac{ou} forward process requires the calculation of normalized Hermite polynomials $\phi_n(x) = \frac{\mathrm{He}_n(x)}{\sqrt{n!}}$. We mainly use the recurrence relation below, for $\ell \geq 1$,
\begin{align}
    \phi_{\ell+1}(x) = \frac{1}{\sqrt{\ell+1}} x\phi_\ell(x) - \sqrt{\frac{\ell}{\ell+1}} \phi_{\ell-1}(x), \qquad \phi_0(x) = 1,  \quad \phi_1(x) = x; \label{eq: hermite recurrence}
\end{align}
see \cite{abramowitz1965handbook}.
The recurrence relation circumvents the explicit expression of $\phi_n(x)$ as a polynomial, which might include large numerical coefficients for some $x^p$. 

As we require the polynomial decomposition of $\phi_k\phi_{\ell}$ in order to evaluate the carr\'{e}-du-champ operator, we further make use of the recurrence relation to obtain a dynamic programming approach to evaluate $\phi_k\phi_\ell = \sum_{h=1}^m \beta_h^{(k, l)} \phi_h$. Firstly, we note that $\mbbeta^{(k, 0)} = \mathbbm{1}^{(k)}$ by definition, where $\mathbbm{1}^{(k)}$ refers to the one-hot vector representation where the $k$-th entry is $1$, with all other entries being $0$. In addition, as $x=\phi_1(x)$, the recurrence relation \eqref{eq: hermite recurrence} implies the following
\begin{align*}
    \mbbeta^{(k, 1)} &= \sqrt{k+1}\mbbeta^{(k+1, 0)} + \sqrt{k} \mbbeta^{(k-1, 0)}.
\end{align*}
Multiplying both sides of equation \eqref{eq: hermite recurrence} with $\phi_k$, we derive the high-order recurrence to calculate $\mbbeta^{(k, \ell)}$ for $\ell \geq 2$:
\begin{align}
    \mbbeta^{(k, \ell+1)} &= \sqrt{\frac{k+1}{\ell+1}}\mbbeta^{(k+1, \ell)}+ \sqrt{\frac{k}{\ell+1}}\mbbeta^{(k-1, \ell)} - \sqrt{\frac{\ell}{\ell+1}} \mbbeta^{(k, \ell-1)}. \label{eq: hermite interaction relation}
\end{align}

%Here we recall a useful property of Hermite polynomials which facilitates higher-order score approximation for the \ac{ou} process as described in \Cref{sec: operator informed}.
%The property that we exploit is a recurrence relation for the probabilist's Hermite polynomials, which is initialized with $\text{He}_0(x) = 1, \text{He}_1(x) = x$, and then follows
%\begin{align*}
%    \text{He}_{n+1}(x) = x \text{He}_n(x) - n \text{He}_{n-1}(x) .
%\end{align*}
%Based on this recurrance relation, it is straight forward to recursively obtain the polynomial coefficients of each $\text{He}_n$, which we denote as $\text{He}_i(x) = \sum_{m=0}^i q_m^{(i)} x^m$. 
%Now, given an arbitrary polynomial $f(x;\mbp) := \sum_{m=0}^n p_m x^m$, the task that we need to address is how to obtain the coefficients $\mbr$ of the Hermite series $f(x;\mbp) = \sum_{m=0}^n r_m \text{He}_m(x)$.
%This task can be solved by observing that $\mbr$ satisfies the linear system
%\begin{align*}
%    \left(\begin{matrix} \mbq^{(0)} & \mbq^{(1)} & \hdots & \mbq^{(n)}
%    \end{matrix}\right) \mbr = \mbp.
%\end{align*}
%This approach can be used to express products $\phi_i \phi_j$ of eigenvalues of the \ac{ou} process as sums of individual $\phi_i$, as explained in the main text. 
%
%
%
%
%
%
%
%
%
%
%

\section{The Special Case of the Brownian Motion Process}
% \mr{This appendix is not referenced in the main text, consider adding a comment around line 203 in the future.}
\label{app: truncated bm}
The Brownian motion 
does not have finite invariant measure,  $\pi$ being the Lebesgue measure. As a consequence, even if its infinitesimal generator has a simple form, its spectral decomposition is less 
% well-defined, 
straightforward,
compared to stable Markov diffusion processes.

\subsection{Spectrum of the \ac{bm} Infinitesimal Generator}
%\textcolor{teal}{[To answer Marina's questions (I should have also expressed it better), even though $\exp(\imath\xi x)$ is an eigenfunction of $\Delta$ by definition, they do not belong to the domain.]}
For any given Markov semigroup and its generator $\mathscr{L}$, a function $f$ belongs to the \emph{domain} $\mathcal{D}(\mathscr{L})$ of the infinitesimal generator if it is square integrable with respect to the invariant measure $\pi$ and it satisfies the following
\begin{align}
    \int \mathscr{L} f(\mbx) \mathrm{d}\pi(\mbx) = 0.
    \label{eq:domain_L}
\end{align}
As $\mathscr{L}$ is conceptually the time derivative of $P_t$, and $f$ should satisfy the invariant condition $\mathbb{E}_{\mbX\sim\pi}[f(\mbX)] = \mathbb{E}_{\mbX\sim\pi}[P_tf(\mbX)]$, we can see that even if a function might satisfy $\mathscr{L} f(x)=\lambda f(x)$ pointwise for some $\lambda \in \mathbb{C}$, it might 
%\textcolor{red}{\sout{not be an eigenfunction as it could}}\mr{I am still keen to call these eigenfunctions, if they satisfy the eigenrelation. Simply they do not belong to the Stein class of the generator, which is the domain $\mathcal{D}(\mathscr{L})$.}
violate square integrability with respect to the Lebesgue measure or the invariant condition \eqref{eq:domain_L}, and thus be excluded from the domain of $\mathscr{L}$. In fact, for the one-dimensional \ac{bm}, functions $\phi_\xi(x)=\exp(\imath \xi x)$ formally have eigenvalue $-\xi^2$, but they do not belong to $\mathcal{D}(\mathscr{L})$ as they are not square integrable and $\int \phi_\xi(x) \mathrm{d}\pi(x)$ do not exist. 

Despite the above caveat, we are free to assume that $\int \phi_\xi\mathrm{d}\rho_0(x)$ does exist, and $P_t\phi_\xi(x)$ can be calculated in closed form:
\begin{align}
    P_t \phi_\xi(x) &= \int \exp(\imath \xi y) \mathcal{N}(y\vert x, 2t)\mathrm{d}y=\exp(-\xi^2 t + \imath \xi x) = \exp(-\xi^2 t) \phi_\xi(x), \label{eq: brownian eigenrelation}
\end{align}
which is a direct result following the characteristic function of a Gaussian distribution. Therefore, the predictable time evolution $\mathbb{E}_{\rho_t}[\phi_\xi] = e^{-\xi^2 t} \mathbb{E}_{\rho_0}[\phi_\xi]$ still holds despite the domain issue. 

Apart from the lack of square integrability, the uncountable nature of the \ac{bm} eigenfunctions poses another issue. In fact, the eigenvalues (which are non-positive) represent a measure of importance of the corresponding eigenvectors, with   larger (closer to 0) eigenvalues indicating a slower decay of $\mathbb{E}_{\mbX_t \sim \rho_t} [ \phi_n(\mbX_t)]$,
as $t$ increases, following \eqref{eq: eigenrelation}. However, because $\phi_\xi$ are uncountable, it is not possible to enumerate the ``top $n$ most important eigenfunctions'', which is  an essential  task for accurately defining the energy-based model used in \ac{oism}. 

\subsection{Truncated \ac{bm} Process}
We propose to address both issues encountered by using the \ac{bm}, by considering a different but related forward process, with 
modified domain and invariant measure. Our reasoning broadly follows that of Chapter 3 of \citet{bakryAnalysisGeometryMarkov2013}, which outlines that (symmetric) Markov diffusion operators could be effectively abstracted into the \emph{Markov triple} $(E, \pi, \Gamma)$, that is, the domain $E$ of $\mbx$, a $\sigma$-finite invariant measure $\pi(\mathrm{d}\mbx)$ and the carr\'e-du-champ operator $\Gamma(\cdot, \cdot)$. Conceptually, given the Markov triple and a suitable function set within the domain of $\Gamma$, it is possible  extrapolate the infinitesimal generator $\mathscr{L}$ and the Markov semigroup $P_t$. 
%\textcolor{red}{Markov diffusion operators exist outside specific instances of diffusion processes, and}\mr{Not sure what this exactly means} 
%one can fully characterize the generator $\mathscr{L}$ and Markov semigroup $P_t$ using a \emph{Markov triple} $(E, \pi, \Gamma)$, that is, the domain $E$ of $\mbx$, a $\sigma$-finite invariant measure $\pi(\mathrm{d}\mbx)$ and the carr\'e-du-champ operator $\Gamma(\cdot, \cdot)$.

For example, the Markov triple of the Brownian motion is formally $(\mathbb{R}, \mathrm{d}x, f'(x)g'(x))$, %\mr{use first derivatives of keep gradients for compactness?} 
but the Lebesgue measure $\mathrm{d}x$ is not $\sigma$-finite. However, if we constrain the domain of $x$ to be a finite interval, chosen as $E=[-\pi, \pi]$, and specify the invariant measure $\pi$ as the uniform distribution on $E$, the resulting Markov triple satisfies all conditions, namely, the $\sigma$-finiteness of the invariant measure.  %\mr{It is not clear which conditions and what are these necessary/sufficient for - see also comment above} 
%\textcolor{teal}{[ZS: Strictly speaking, we have only just proposed the Markov triple, and how to use it to extrapolate the form of $P_t$ or sample from it remains unknown. My explanation is mainly by construction that illustrates that the 2-step approach of sampling from BM + constraining to interval satisfies all conditions.]} \mr{Ok - if you could explain it, even in plain words that would be good. Because in practice we might use this truncated BM ? }
%\textcolor{teal}{[ZS: I think in future work, we should explore the possibility of OISM+NN with truncated BM as forward process -- the DSM objective would actually change.]}

Strictly speaking, this altered Markov triple induces a different generator and semigroup, and by extension, a completely new forward process, which we call \emph{truncated Brownian motion}.
%\mr{it seems that this is called \emph{wrapped Brownian motion} or \emph{Brownian motion on the circle}, an example of BM on manifolds, see `Stochastic Analysis On Manifolds' Elton P. Hsu, example 3.3. The transition density given below is the heath kernel on the circle, therefore converges to the uniform distribution. Is it fair to say that the generator behaves in the same way as that of the BM, because of it connection to the semigroup, and the fact that the processes are locally undistinguishable, but their domains are different, because when using the wrapped BM we naturally consider periodic functions}. 
%\textcolor{teal}{[ZS: I think BM on a circle (it starts from 1D Wiener process, but extends to a 2D circle) is not exactly the same. Our version is even simpler that it is just 1D constrained on an interval. We can add ``locally indistinguishable'' as a reason for the explanation below.]} \mr{ok that sounds good}
We demonstrate that this forward process largely coincides with Brownian motion with one single revision. Our description focuses on the 1-dimensional setting, but our argument extends trivially to high-dimensions. 
\begin{example}[Truncated \ac{bm}]
    To sample $X_t$ from a truncated \ac{bm}, given the initial state $X_0=x$, we first sample $\tilde{X}_t\sim\cN(x, 2t)$ according to the Brownian motion \eqref{eq: bm process}, and then shift $\tilde{X}_t$ using multiples of $2\pi$ until it falls into the interval $[-\pi, \pi]$. Specifically,
    \begin{align*}
        X_t = \begin{cases}
            \tilde{X}_t \bmod 2\pi, &\qquad \text{if } \left(\tilde{X}_t \bmod 2\pi\right) \leq \pi,\\
            \tilde{X}_t \bmod 2\pi - 2\pi, &\qquad \text{otherwise.}
        \end{cases}
    \end{align*}
\end{example}
Therefore, the conditional distribution $\rho_t(y\vert x)$ follows an infinite sum
\begin{align}
    \rho_t(y\vert x) = \sum_{z\in\mathbb{Z}} \cN(y\vert x +2\pi z, 2t). \label{eq: truncated rho_t}
\end{align}
We could see that, as $t\rightarrow\infty$, the conditional distribution converges to a uniform distribution, which is therefore the stationary distribution. %\textcolor{teal}{[ZS: this looks trivially true but I'm not actually sure how to explain it.]} 
Since a truncated \ac{bm} is indistinguishable from an unconstrained \ac{bm} given a small $t$, the generator $\mathscr{L}$ and the carr\'{e}-du-champ operator coincides with those of the Brownian motion. %\textcolor{teal}{[ZS: also a trivially true but hard to explain statement.]}
Moreover, the functions $\left\{\sqrt{2}\exp(\imath n x)\right\}_{n\in\mathbb{Z}}$ form an orthonormal basis  on $L^2([-\pi, \pi])$,  and they have period $2\pi$ when extended onto the real line $\mathbb{R}$ and therefore satisfy \eqref{eq:domain_L}. Furthermore, 
\begin{align*}
    P_t\phi_n(x) &= \sum_{z\in\mathbb{Z}}\int_{-\pi}^\pi \exp(\imath n y)\mathcal{N}(y\vert x + 2\pi z, 2t)\mathrm{d}y \\
    &= \int_{\mathbb{R}} \exp(\imath n y) \mathcal{N}(y\vert x, 2t)\mathrm{d}y=\exp(-n^2 t + \imath n x) = \exp(-n^2 t) \phi_n(x).
\end{align*}
Most notably, in the truncated setting \eqref{eq: truncated rho_t}, the eigen-relation \eqref{eq: brownian eigenrelation} does not apply to all functions $\phi_\xi =\exp(\imath\xi x)$, as re-writing the infinite sum into an integral on $\mathbb{R}$ in the second identity above applies to only functions with a period of $2\pi$, adding the additional constraint of $\xi \in \mathbb{Z}$. Therefore, the eigen-pairs contained within $\mathcal{D}(\mathscr{L})$ only include $\left(-n^2, \sqrt{2}\exp(\imath n x)\right)_{n\in\mathbb{Z}}$, and the eigenfunctions satisfy both square integrability with respect to $\pi$ and the invariance property. 

As a consequence, by constraining the \ac{bm} on a finite interval, we obtain a diffusion process with a finite stationary measure, as well as a better-defined countable set of eigenvectors, which can be sorted by importance. 
%The \emph{truncated} variant of has a countable spectrum $\left(-n^2, \sqrt{2}\exp(\imath n x)\right)_{n\in\mathbb{Z}}$, as the trigonometric functions become square integrable on a finite interval. 
%Functions $\phi_\xi$ with non-integer $\xi$s do not satisfy $\int \mathscr{L} \phi_\xi \mathrm{d}\pi = 0$, and the eigen-relation \eqref{eq: brownian eigenrelation} no longer holds on the constrained interval 
%\begin{align*}
%    P_t \phi_\xi(x) = \int_{-\pi}^\pi \exp(\imath\xi y) \mathcal{N}(y|x, 2t)\mathrm{d}y \ne \exp(-\xi^2 t)\phi_\xi(x).
%\end{align*}
%and thus do not belong in $\mathcal{D}(\mathscr{L})$. 
%\mr{Not belonging to $\mathcal{D}(\mathscr{L})$ did not prevent the BM eigenfunctions to be defined as functions in its spectrum, so I am not sure why would this prevent the $ \phi_\xi$ from being in the spectrum of the truncated BM} 
%Given that the spectrum of $\mathscr{L}$ contains only of periodic functions, \mr{this implication is not clear to me} we note that forward sampling from the truncated Brownian forward process involves first simulating a Gaussian sample according to the unconstrained Brownian motion, \mr{do you refer to the conditional distribution here?} and then shift the sample with multiples of $2\pi$ such that it falls into the interval $[-\pi, \pi]$. 

\paragraph{Enumerating the eigenfunctions} The truncated Brownian motion allows for the enumeration of the ``top-$n$ most important eigenfunctions''. In order to circumvent working with complex numbers, we make the practical choice to use the sine and cosine functions instead of complex exponentials, that is, we consider the eigenpairs 
\begin{align*}
    \left\{
\left(-k^2, \sqrt{2}\cos(kx)\right)\right\}_{k\in[n]}\bigcup \left\{
\left(-k^2, \sqrt{2}\sin(kx)\right)\right\}_{k\in[n]}
\end{align*}
for the spectrum of the generator $\mathscr{L}$. The representation of the product of two trigonometric as sums
% of other trigonometric functions functions 
can be obtained via simple trigonometric identities. In the multi-dimensional setting, we first set up a threshold of the minimum eigenvalue $-n_d$, and denote $\mbXi(n_d) := \{\mbxi\in\mathbb{Z}^d: \norm{\mbxi}^2\leq n_d\}$. Then the set of eigenpairs consists of 
\begin{align*}
    \left\{\left(-\norm{\mbxi}^2, \sqrt{2}\cos(\mbxi^\top\mbx)\right)\right\}_{\mbxi\in\mbXi(n_d)}\bigcup \left\{\left(-\norm{\mbxi}^2, \sqrt{2}\sin(\mbxi^\top\mbx)\right)\right\}_{\mbxi\in\mbXi(n_d)}.
\end{align*}
%\begin{align}
%    \forall n \in \mathbb{Z}, \int \mathscr{L}  \mathrm{d}\pi(x) = 0.
%\end{align}

\section{Additional Notes on \ac{oism}}
\label{app: oism addition}
This section outlines the usage of orthogonal functions in density estimation and its connection to Stein's normal means problem in \Cref{app: orthogonal series}, and the specific form of shrinkage estimators in \Cref{app: modulation}, and a simple procedure for a preconditioned linear solver in \Cref{app: preconditioner}.

\subsection{Orthogonal Series Density Estimation}
\label{app: orthogonal series}
Orthogonal functions in density estimation was first proposed by \citet{whittleSmoothingProbabilityDensity1958, cencov1962estimation, schwartzEstimationProbabilityDensity1967}, and it remains a topic of interest, see the recent works in variational inference \citep{cai2024eigenvi}.
Given a set of basis functions $\left(\phi_n\right)_{n\in\mathbb{N}}$ in the space of square-integrable functions $L^2(\pi)$ and assuming the density ratio $\frac{\rho}{\pi}$ is square-integrable, we can decompose the density ratio into the following
\begin{align}
    \frac{\rho}{\pi}(\mbx) &= \sum_{n\in\mathbb{N}} \mathbb{E}_{\rho}\left[\phi_n\right] \cdot \phi_n(\mbx) \approx \sum_{n\in\mathbb{N}} \left(\frac{1}{M} \sum_{m=1}^M \phi_n(\mbx_m)\right) \cdot \phi_n(\mbx) =: \frac{\widehat{\rho}}{\pi}(\mbx).\label{eq: density ratio}
\end{align}
The right-hand side of \eqref{eq: density ratio} refers to the infinite series with sample means plugged in for each coefficient, which is typically numerically unstable, or even not exist at all. For example, in the case of Hermite polynomials, the infinite series holds in the sense of distributions; see Section II.9 of \citet{courant1954methods}
\begin{align*}
    \sum_{n\in\mathbb{N}} \phi_n(\mbx) \phi_n(\mbx_0) = \delta(\mbx-\mbx_0). 
\end{align*}
Therefore, the right-hand side of the infinite series yields the density estimate $\frac{1}{M} \sum_{m=1}^M \delta(\mbx-\mbx_m)$, indicating the superfluous ``density estimate'' that simply involves Dirac delta's on the $M$ samples themselves. This is an undesirable outcome, as it defeats the purpose of density estimation and generative modeling:  a model that generates exact copies from within the dataset $\{\mbx_m\}_{m=1}^M$ lacks the ability to generate novel, unseen samples. 

The infeasibility of replacing the orthogonal series coefficients with sample averages \eqref{eq: density ratio} could also be seen from the inadmissibility of the sample mean estimator in normal means estimation \citep{steinInadmissibilityUsualEstimator1956}. Employing the central limit theorem, we could derive that when $\rho$ and $\pi$ are sufficiently similar, the sample means estimator $\widehat{\mbtheta}\in \mathbb{R}^n$ with components $\widehat{\mbtheta}^{(k)} := \frac{1}{M}\sum_{m=1}^M \phi_k(\mbx_m)$ %\mr{I am confused by the notation $\widehat{\mbtheta}_k $ and the bold notation, here and below. Ah now I see, with $\widehat{\mbtheta}_k$ you meant the k-th component of the vector $\widehat{\mbtheta}$. But previously in the paper we would have used $\widehat{\mbtheta}^{(k)}$. Might need to change below too}
%\textcolor{teal}{[ZS: I will make the changes]}
converges in distribution to the Gaussian $\cN\left(\mbtheta, \frac{1}{M}\mbSigma\right)$, where
\begin{align*}
    \mbtheta^{(k)} := \mathbb{E}_{\rho}[\phi_k], \qquad \mbSigma^{(k,\ell)} := \mathbb{E}_{\rho}\left[\left(\phi_k-\mbtheta^{(k)}\right)\left(\phi_\ell-\mbtheta^{(\ell)}\right)\right] \approx \delta_{k\ell},
\end{align*}
thus recovering the exact same scenario as Stein's paradox. In fact, when orthogonal functions are used for nonparametric regression and density estimation, the need for lower-risk shrinkage estimators has been widely acknowledged; see Chapter 8 of \citet{wassermanAllNonparametricStatistics2005} and \citet{efromovichOrthogonalSeriesDensity2010} for detailed background review on suitable choices for linear shrinkage. 
\subsection{Modulation Estimator}
\label{app: modulation}
We explain in this section how we implement the linear shrinkage estimator. Given the sample means estimator $\widehat{\mbtheta}$, the simplest linear shrinkage would be the James-Stein estimator \citep{james1961estimation}, which uses the estimator $\gamma \widehat{\mbtheta}$, where $0\leq \gamma \leq 1$ is a scalar-valued shrinkage parameter. We opt for the slightly more flexible \emph{modulation estimator} \citep{beranModulationEstimatorsConfidence1998}, which uses $\mbgamma \circ \widehat{\mbtheta}$, where $\mbgamma \in [0, 1]^{n}$ is a vector-valued shrinkage estimator, and $\circ$ denotes Hadamard (element-wise) product. 

To construct the modulation estimator, we require an estimate of the variance of the sample mean, which is obtained as follows
\begin{align*}
\hat{\sigma}_k^2 = \frac{1}{M^2} \sum_{m=1}^M\left(\phi_k(\mbx_m) - \widehat{\mbtheta}^{(k)}\right)^2.
\end{align*}
Then the 
empirical risk is obtained as 
\begin{align}
    \widehat{R}(\mbgamma) = \sum_{k=1}^n \gamma_k^2 \hat{\sigma}_k^2 + (1-\gamma_k)^2\max\left\{\left(\widehat{\mbtheta}^{(k)}\right)^2-\hat{\sigma}_k^2, 0\right\}, \label{eq: modulation risk}
\end{align}
which is minimized at 
 $\widehat{\mbgamma} = \arg\min_{\mbgamma\in[0,1]^n} \widehat{R}(\mbgamma)$.
\subsection{Preconditioned Linear Solver}
\label{app: preconditioner}
\ac{oism} requires solving the linear system $\mbA_t\widehat{\mbalpha} = \mbb_t$. To enhance numerical stability, we adopt a simple preconditioner, based on the fact that $\{\phi_n\}$ are orthonormal and $\mathbb{E}_{\pi}[\mathscr{L}(\phi_k\phi_{\ell})] = 0$.
\begin{align*}
\int \Gamma(\phi_k(\mbx), \phi_\ell(\mbx))\mathrm{d}\pi(\mbx) &= \frac{1}{2} \int \mathscr{L}(\phi_k\phi_\ell)(\mbx)\mathrm{d}\pi(\mbx) - \frac{\lambda_k+\lambda_\ell}{2}\int \phi_k(\mbx)\phi_\ell(\mbx)\mathrm{d}\pi(\mbx) 
\\
&= -\frac{\lambda_k+\lambda_\ell}{2} \delta_{k\ell},
\end{align*}
%\textcolor{red}{given that we work with eigenfunctions that lie in the domain of the generator, and whose product is in the same class  (Hermite polynomials or trigonometric functions).}
As $\rho_t$ converges to $\pi$ as $t\rightarrow\infty$, the matrix $\mbA_t$ converges to a diagonal matrix $\mbLambda:=\mathrm{diag}(-\lambda_1, -\lambda_2, \hdots, -\lambda_n)$. Therefore, we opt to solve the \emph{preconditioned} system $\mbLambda^{-1}\mbA_t\widehat{\mbalpha} = \mbLambda^{-1}\mbb_t$.

\section{Further Consequences of Markov Diffusion Operators}
\label{app: consequences}
%\textcolor{teal}{[ZS: I think this section does not propose new methodology, mainly to showcase that derivations can be simplified with the help of diffusion operators. ]}
The aim of this appendix is to revisit objectives related to score matching through the lens of Markov diffusion operators, specifically 
%\emph{Tweedie's formula} \citep{tweedieFunctionsStatisticalVariate1947}, 
\emph{time score matching} \citep{choiDensityRatioEstimation2022}, the \emph{score} \acf{fpe} \citep{laiImprovingScorebasedDiffusion2023}, and \emph{generalized} score matching \citep{bentonDenoisingDiffusionsDenoising2024}. 
%\textcolor{teal}{[ZS: I think the note on Tweedie's formula is not interesting, and it would be nice if we add a note explaining the Benton paper.]}\mr{In line 189 we mentioned we would talk about Tweedie's formula}
Our main tool of analysis is the \emph{diffusion property} of the generator $\mathscr{L}$ which states that, for sufficiently smooth $f$ and $\psi: \mathbb{R} \rightarrow \mathbb{R}$, the following change of variable formula applies
\begin{align}
    \mathscr{L}\psi(f) &= \psi'(f)\mathscr{L} f + \psi''(f)\Gamma(f, f); \label{eq: diffusion property}
\end{align}
see Definition 1.11.1 of \citet{bakryAnalysisGeometryMarkov2013}.

\paragraph{Insight into time score matching}
An alternative to the score matching objective is to match the time derivative of the log density ratio, an approach termed \emph{time score matching} in \citet{choiDensityRatioEstimation2022}, which, analogous to score matching, minimizes the expected squared loss with respect to the \emph{temporal} derivative
%Here we point out that time score matching can be formulated in an analogous manner to \eqref{eq: score matching p0} using the Markov semigroup, based on the identity 
\begin{align*}
    \mathbb{E}_{\mbX_t \sim \rho_t} \left[ \frac{\partial}{\partial t} f_t(\mbX_t) - \frac{\partial}{\partial t}\log\frac{\rho_t}{\pi}(\mbX_t)\right]^2 
    \stackrel{+C}{=} &\mathbb{E}_{\mbX_0 \sim\rho_0} \left[P_t \left(\frac{\partial}{\partial t} f_t\right)^2(\mbX_0)\right]\\
    & \,\,\, -  2\mathbb{E}_{\mbX_t\sim\rho_t}\left[ \frac{\partial}{\partial t} f_t(\mbX_t) \right] \left[ \frac{\partial}{\partial t} \log\frac{\rho_t}{\pi}(\mbX_t) \right].
    %& \stackrel{+C}{=} \mathbb{E}_{\mbX_0 \sim\rho_0} \left[P_t \left(\left(\frac{\partial}{\partial t} f_t(\mbX_0)\right)^2 - 2\mathscr{L} \frac{\partial}{\partial t} f_t(\mbX_0)\right)\right].
\end{align*}
%\mr{the right-hand side is  ambiguous, $P_t$ should apply to a function}
%The above formula is obtained by exploiting self-adjointness of $\mathscr{L}$ \textcolor{red}{when we work with symmetric semigroups}\mr{I don't think we have defined self-adjointness before}. Namely,
The intractable term can similarly be resolved via integration by parts, and the time evolution of density ratio \eqref{eq: operator fpe}
\begin{align*}
     \int \left[ \frac{\partial}{\partial t} f_t(\mbx) \right] \left[ \frac{\partial}{\partial t} \log\frac{\rho_t}{\pi}(\mbx) \right] \d{\rho_t(\mbx)} = & \int \left[ \frac{\partial}{\partial t} f_t(\mbx) \right] \left[ \frac{\partial}{\partial t}\frac{\rho_t}{\pi}(\mbx) \right] \d{\pi(\mbx)}\\
    =& \int \left[ \frac{\partial}{\partial t} f_t(\mbx) \right] \mathscr{L}  \frac{\rho_t}{\pi}(\mbx)  \d{\pi(\mbx)} = \int \mathscr{L} \left( \frac{\partial}{\partial t} f_t\right)(\mbx)  \d{\rho_t(\mbx)} \\
    =& \mathbb{E}_{\mbX_0\sim\rho_0}\left[P_t\mathscr{L} \left( \frac{\partial}{\partial t} f_t\right)(\mbX_0)\right].
\end{align*}
Therefore, matching the time derivative similarly offers a perspective analogous to the implicit score matching derived via \ac{oism} \eqref{eq: oism main}.
Time score matching and score matching optimize the temporal and spatial partial derivatives of the forward process, respectively. We illustrate that the process of removing intractability from the original mean squared error form follows a similar pattern. We also obtain a notably different solution compared to \citet{choiDensityRatioEstimation2022}, which does not make use of the density ratio \ac{pde} \eqref{eq: operator fpe}.

\paragraph{Generalizing the score Fokker--Planck equation}

Sharing a similar starting point to this work, \citet{laiImprovingScorebasedDiffusion2023} noted that traditional score matching is agnostic 
%\mr{do you mean `influenced by'/`affected by' here?} 
to the specific type of the forward process.
The authors reasoned that the true score function ought to satisfy a \ac{pde} and derived an explicit \ac{pde} in the case of a Markov diffusion forward process.
Here we present a streamlined and more general derivation of that result.
The diffusion property \eqref{eq: diffusion property} enables us to study the time evolution of the density ratio $\rho_t / \pi$, and we have that
\begin{align}
    \frac{\partial}{\partial t}\log \frac{\rho_t}{\pi} (\mbx) = \frac{\frac{\partial}{\partial t} P_t \frac{\rho_0}{\pi}(\mbx)}{\frac{\rho_t}{\pi} (\mbx)} = \log'\left(\frac{\rho_t}{\pi} (\mbx)\right) \mathscr{L} \frac{\rho_t}{\pi} (\mbx) = \mathscr{L}\log \frac{\rho_t}{\pi} (\mbx) + \norm{\nabla_\mbx \log \frac{\rho_t}{\pi} (\mbx)}^2 \label{eq: time deriv formula}
\end{align}
where the final equality follows from \eqref{eq: diffusion property} with $\psi(f) = \log(f)$ and the explicit form of the carr\'{e}-du-champ operator for the \ac{ou} process in \Cref{ex: operators for OU}.
The time derivative of the true score function $\mbs_t^\star(\mbx) :
= \nabla\log\frac{\rho_t}{\pi}(\mbx)$ 
%\mr{define $\mbs_t^\star(\mbx)$, we have not used this notation before} 
in \eqref{eq: true score} can be obtained by taking the $\mbx$ derivative of \eqref{eq: time deriv formula}, leading in the case of the \ac{ou} process to the \ac{pde}
\begin{align}
    \frac{\partial}{\partial t} \mbs_t^\star(\mbx) = \nabla_\mbx \big[\langle -\mbx, \mbs_t^\star(\mbx)\rangle + \mathrm{tr}(\mbJ(\mbs_t^\star)(\mbx))\big] + 2 \mbJ(\mbs_t^\star)(\mbx) \mbs_t^\star(\mbx) \label{eq: score FP}
\end{align}
where $\mbJ(\mbs_t^\star)$ denotes the Jacobean of $\mbs_t^\star$, which recovers the \emph{score Fokker--Planck equation} as coined by \citet{laiImprovingScorebasedDiffusion2023}. 
In that work the authors noted that standard score matching typically violates \eqref{eq: score FP}, especially for small values of $t$ where the regularity of $\rho_t / \pi$ is most pronounced \citep[c.f. Fig. 1 in][]{laiImprovingScorebasedDiffusion2023}.
Based on this observation, the authors proposed to add a regularization term to the score matching objective that aims to explicitly enforce \eqref{eq: score FP} to (approximately) hold.
The derivation that we have presented generalizes beyond the \ac{ou} process by substituting alternative expressions for the infinitesimal generator and the carr\'{e}-du-champ operator.

\paragraph{Generalized score matching} \citet{bentonDenoisingDiffusionsDenoising2024} propose a \emph{generalized} score matching objective that does not include a $L^2$ norm and can thus be used for non-euclidean data. The intuition behind it relates to the \emph{diffusion} property \eqref{eq: diffusion property}. With $\psi(f) = \log f$, we can rewrite \eqref{eq: diffusion property} as 
\begin{align*}
    \Gamma(\log f) &= \frac{\mathscr{L}f}{f} - \mathscr{L}\log f.
\end{align*}
Therefore, as we match the score function of $\rho_0$ and an energy-based model $\nu(\mbx)=\exp(-\varphi(\mbx)) / Z$, rewriting the carr\'{e}-du-champ operator yields the generalized form of \emph{explicit} score matching
\begin{align*}
    \mathcal{J}_{\mathrm{ESM}} := \mathbb{E}_{\rho_0}\left[\Gamma(\log\rho_0/\nu)\right] = \mathbb{E}_{\rho_0}\left[\frac{\mathscr{L}\frac{\rho_0}{\nu}}{\frac{\rho_0}{\nu}} - \mathscr{L}\log \frac{\rho_0}{\nu}\right].
\end{align*}
Using the diffusion property, the implicit score matching $\mathcal{J}_{\mathrm{ISM}}$ proposed by \citet{bentonDenoisingDiffusionsDenoising2024} is equivalent to our implicit score matching \eqref{eq: hyvarinen score}, 
\begin{align*}
    \text{(9)} = \mathbb{E}_{\rho_0}\left[\Gamma(\log\nu) + 2\mathscr{L}\log\nu\right] = \mathbb{E}_{\rho_0}\left[\frac{\mathscr{L}\nu}{\nu}+\mathscr{L}\log\nu\right] = : \mathcal{J}_{\mathrm{ISM}}.
\end{align*}
\citet{bentonDenoisingDiffusionsDenoising2024} focus on the appeal of generator-only formula for score matching, and apply it to diffusion models defined on manifolds. 
%\textcolor{teal}{[ZS: There is also a derivation about denoising score matching, but the derivation does not require $\Gamma$ operators so it cannot be simplified. ]}
%
%
%
%
%
%
%
%

%
%
%
%
%
%
%
%
%
%

%
%
%
%

%
%
%
 %
%
%

\section{Experimental Settings: Low-Dimensional Illustrations}
\label{app: experiment}
%\textcolor{blue}{[Zheyang to update this appendix.  Make sure all learning rates, batch sizes, etc are reported]}

This appendix contains full details required to reproduce the one- and two-dimensional experiments that we reported in the main text.

\paragraph{Computational resources}
This experiment was implemented in JAX and performed on the CPU of an Apple MacBook M2 laptop. \ac{oism} takes virtually no training as we only need to compute the shrinkage estimator of the expected eigenfunctions, so the main computational cost comes from density estimation and data generation. 
%The methods take a similar amount of computational resource to run: it is easier for \ac{oism} to perform well in this instance, meaning that a smaller number of iterations were required. 
%
\paragraph{Forward process and selection of eigenfunctions} 
We use the truncated \ac{bm} as the forward noising process, as explained in \Cref{app: truncated bm}. And we use the forward process with a changed noise schedule, such that $0\leq \tau \leq 1$, $\sigma_{\min} = 0.01$, $\sigma_{\max} = 50$. 
\begin{align*}
    \mathrm{d}\mbX_\tau = \sigma_{\min}\left(\frac{\sigma_{\max}}{\sigma_{\min}}\right)^\tau\sqrt{2\log\frac{\sigma_{\max}}{\sigma_{\min}}} \mathrm{d} \mbW_\tau, \qquad \mbX_0 \sim \rho_0.
\end{align*}
For the 1-dimensional example, we choose a combined total of $50$ eigenfunctions, that is $\sqrt{2}\cos(kx)$ and $\sqrt{2}\sin(kx)$ for $k=1, 2, \hdots, 25$. As we require calculation of $\mathbb{E}_{\rho_0}[\phi_\ell]$ on an extended set to account for the carr\'{e}-du-champ operator, the extended set involves $k=1, 2, \hdots, 50$. 

For the 2-dimensional example, we enumerate all eigenfunctions $(\lambda_n, \phi_n)$ such that $\lambda_n \geq -125$, which includes $200$ eigenfunctions. Calculating the carr\'{e}-du-champ operator requires eigenfunctions of eigenvalue no less than $-500$, which includes $790$ eigenfunctions. After calculating the sample mean, we apply the dimension-wise linear shrinkage estimator \eqref{eq: modulation risk}, as explained in \Cref{app: modulation}. 
%Details for standard \ac{dm} are contained in \Cref{subsec: standard DM implement} and details for \ac{oism} are contained in \Cref{subsec: OISM implement}.

\begin{figure}
    \centering
    \includegraphics[width=0.6\linewidth]{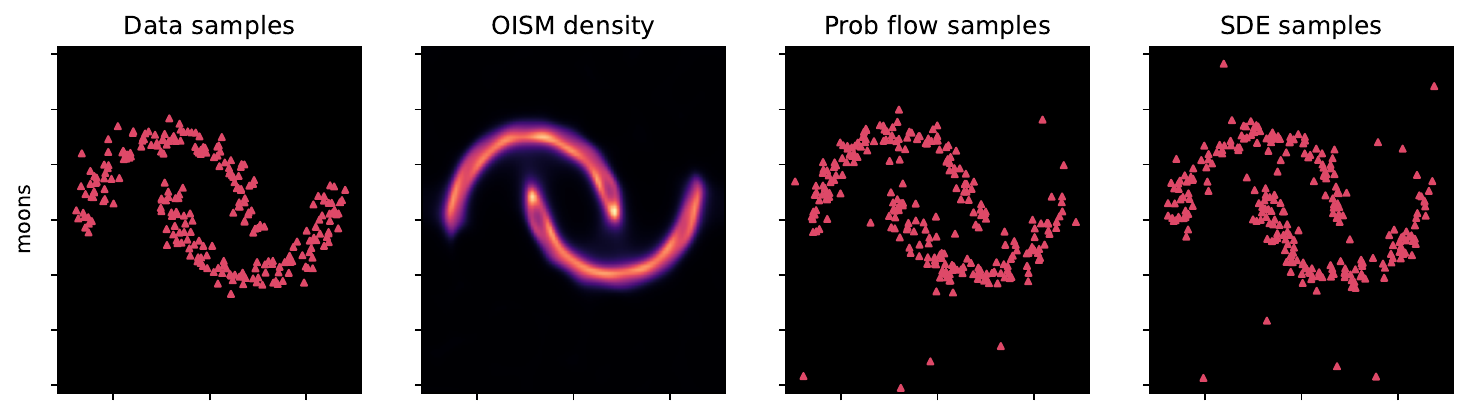}\\
    \includegraphics[width=0.6\linewidth]{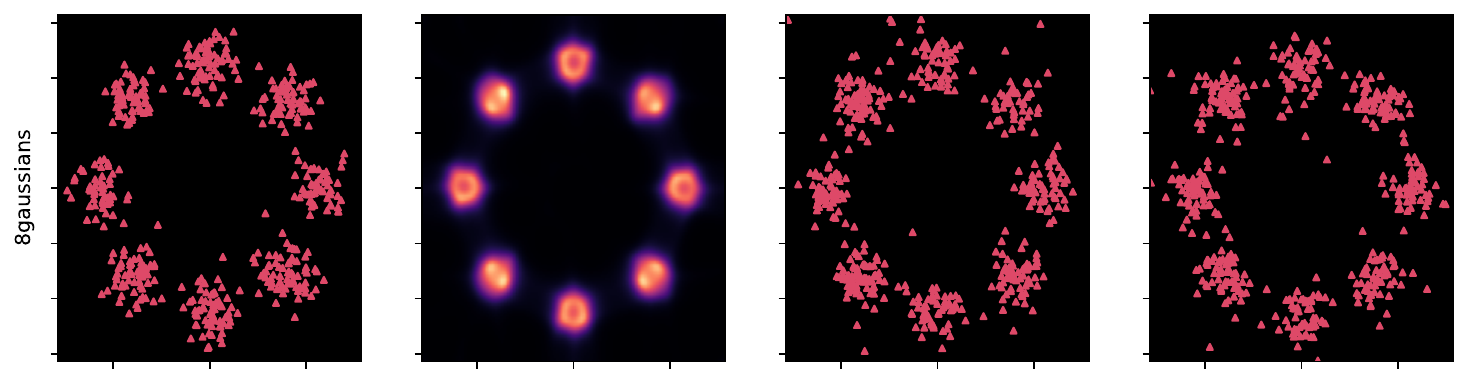}\\
    \includegraphics[width=0.6\linewidth]{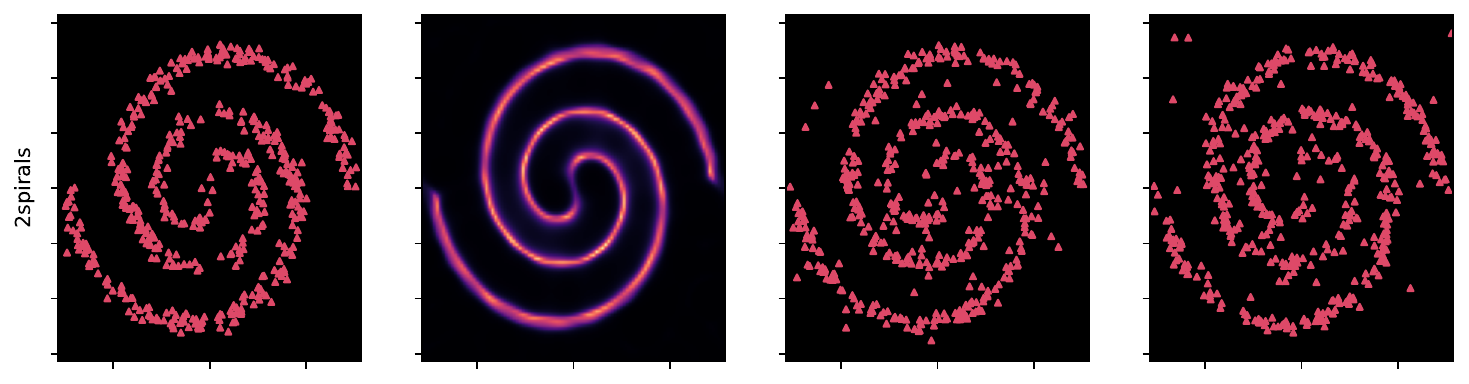}\\
    \includegraphics[width=0.6\linewidth]{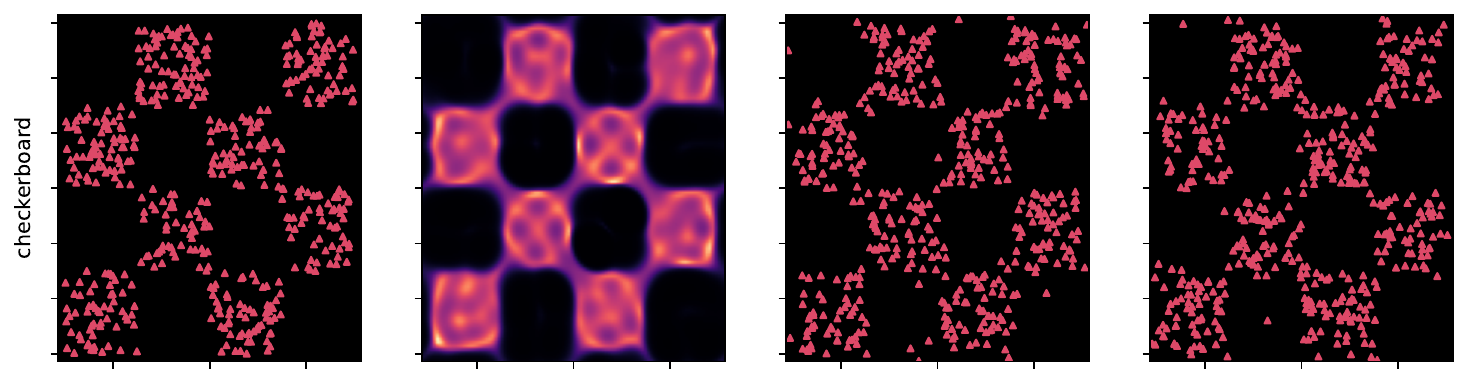}\\
    \includegraphics[width=0.6\linewidth]{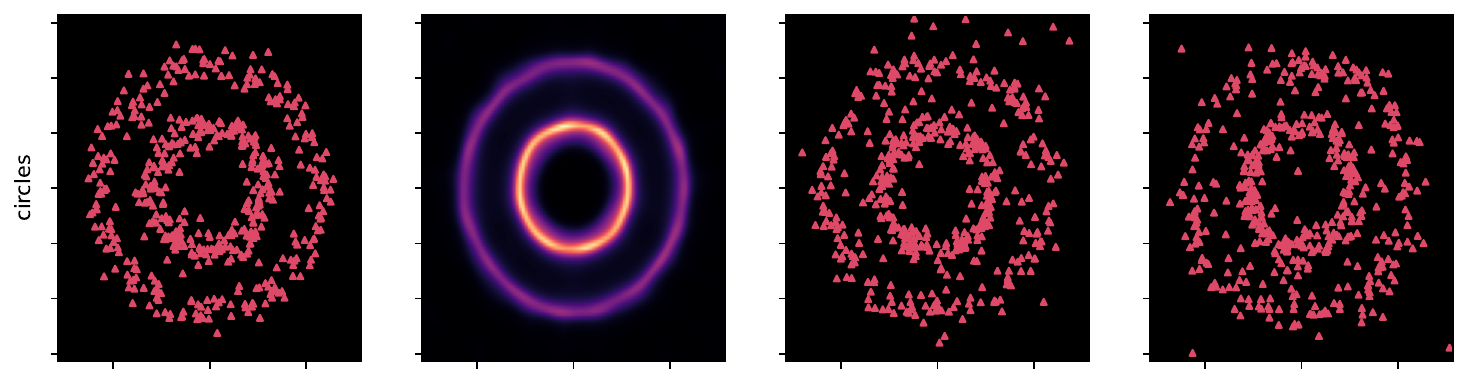}\\
    \includegraphics[width=0.6\linewidth]{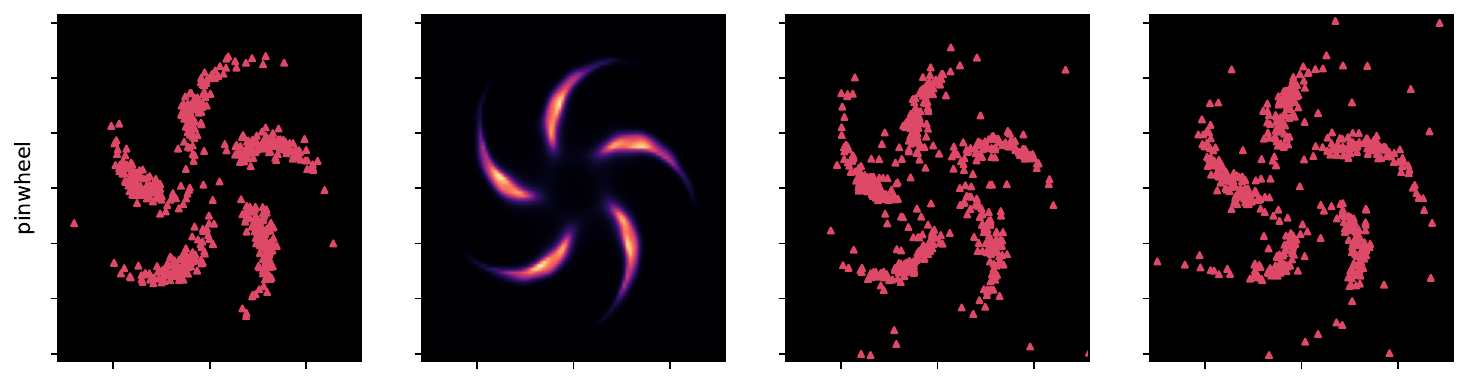}\\
    \includegraphics[width=0.6\linewidth]{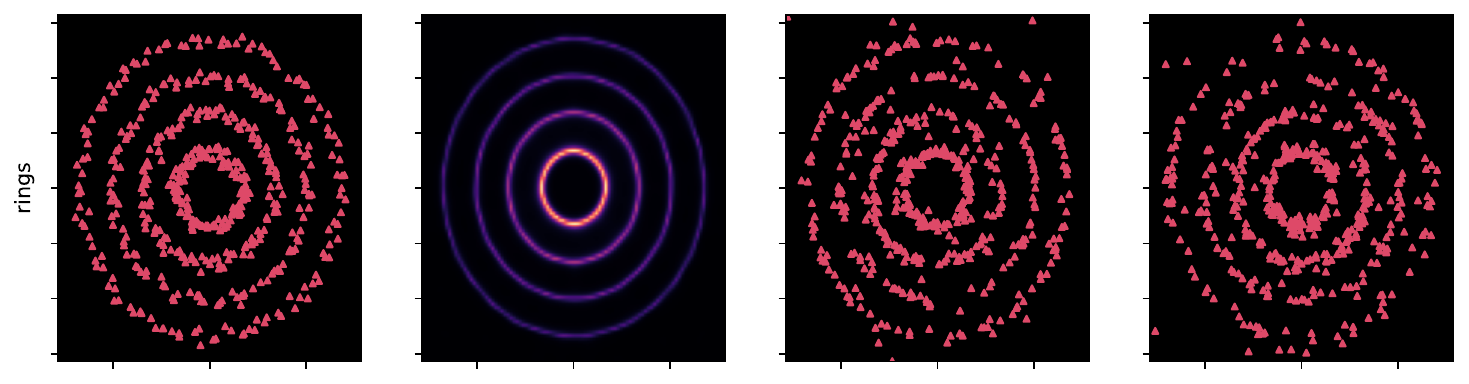}\\
    \includegraphics[width=0.6\linewidth]{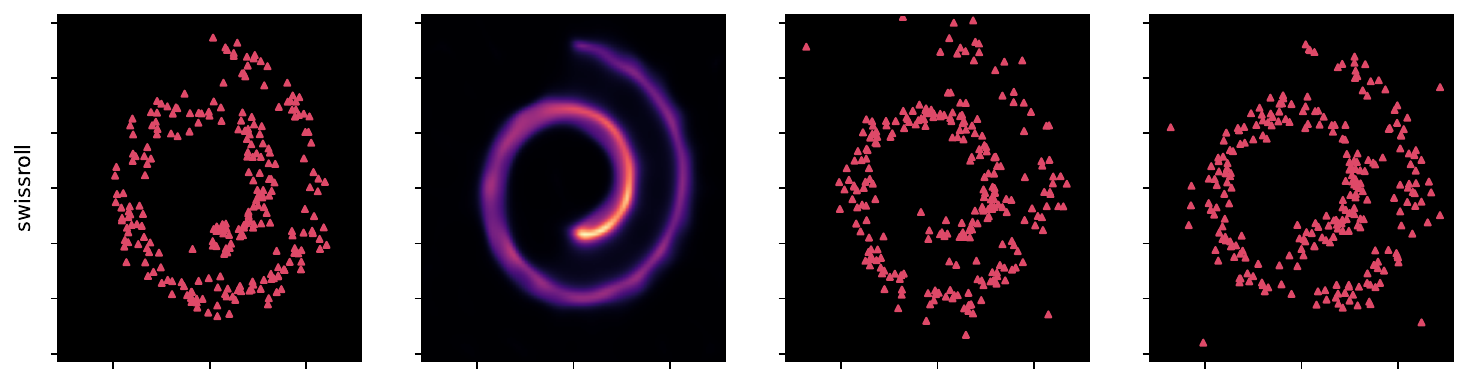}
    \caption{Additional illustrations on toy dataset. The first column shows (a subset of) ground truth samples from the data distribution; The second column visualizes the 
    \ac{oism} density obtained via probability flow \ac{ode}; The third column plots sample points simulated via probability flow \ac{ode}; The last column plots samples generated by a reverse \ac{sde} sampler. }
    \label{fig: toy app}
\end{figure}

\paragraph{Synthetic datasets} 
The Bart Simpson example, described in Chapter 6 of \citet{wassermanAllNonparametricStatistics2005}, is a Gaussian mixture distribution $\rho_0(x) = \frac{1}{2}\cN(x; 0, 1) + \frac{1}{10}\sum_{j=0}^4 \cN(x; j/2-1, 1/100)$, which yields a closed-form marginal distribution $\rho_\tau(x) = \frac{1}{2}\cN(x; 0, 1+\sigma_\tau^2) + \frac{1}{10}\sum_{j=0}^4 \cN(x; j/2-1, 1/100+\sigma_\tau^2)$, where $\sigma_\tau:=\sigma_{\min}\left(\frac{\sigma_{\max}}{\sigma_{\min}}\right)^\tau$. The expected values $\mathbb{E}_{\rho_0}[\phi_k]$ can be calculated in closed form given the characteristic function of Gaussian distributions. A total of $N=2,000$ data points are simulated from the Bart Simpson example. The 2-dimensional synthetic data were generated using existing code that was downloaded from
\url{https://github.com/rtqichen/ffjord} \citep{grathwohlFFJORDFreeFormContinuous2018}, under the MIT license. 
For all experiments, we generated synthetic datasets of size $N = 20,000$. \Cref{fig: toy app} demonstrates similar results obtained via \ac{oism} on a wider range of datasets, as well as samples obtained via simulation of the reverse \ac{sde}. 
\paragraph{Simulation of the probability flow \ac{ode}} In order to generate samples via probability flow ODE~\eqref{eq: prob flow ode}, we need to initialize $\mbX_1\sim \cN\left(\mathbf{0}, \sigma_{\max}^2\mbI\right)$, and rescale $\mbX_1$ within the interval $[-\pi, \pi]$. It is also possible to directly sample $\mbX_1$ from  the invariant measure $\mathrm{Uniform}[-\pi, \pi]$. The exact form is 
\begin{align}
    \frac{\mathrm{d}\mbX_\tau}{\mathrm{d}\tau} = -\sigma_{\min}^2\left(\frac{\sigma_{\max}}{\sigma_{\min}}\right)^{2\tau}\log\frac{\sigma_{\max}}{\sigma_{\min}} \tilde{\mbs}_{t(\tau)}(\mbX_\tau) := \mbf_\tau(\mbX_\tau), \label{eq: prob flow bm}
\end{align}
where $\tilde{\mbs}_{t(\tau)}(\mbx)$ is the \ac{oism} solution \eqref{eq: oism final}, solved via a preconditioner as described in \Cref{app: preconditioner}, for the correct noise level, which corresponds to $t(\tau)=\frac{1}{2}\sigma_{\min}^2\left(\frac{\sigma_{\max}}{\sigma_{\min}}\right)^{2\tau}$. The generation of samples is achieved via first sampling from $\mbX_1 \sim \rho_1$ (the invariant measure) and then simulating \eqref{eq: prob flow bm} backwards from $\tau=1$ to $0$;
if the end-point $\mbX_0$ is outside the range $[-\pi, \pi]$, we need to rescale $\mbX_0$ in the same interval. 
Simulating \eqref{eq: prob flow bm} forward from $\tau=0$ yields a density estimate via the infinitesimal change-of-variables formula \citep{chenNeuralOrdinaryDifferential2018}: $\log \rho_0^{(\text{ODE})}(\mbx_0) = \log \pi(\mbx_1) + \int_0^1 \nabla \cdot \mbf_\tau(\mbx_\tau)\mathrm{d}\tau$, where $\log\pi(\mbx_1)=-d\log2\pi$. We use the adaptive \ac{ode} solver \texttt{tsit5} for both tasks. 

\section{Experimental Settings:  Image Generation}
\label{app: mnist}
%\textcolor{blue}{[Huihui and Zheyang to add/update the details in this appendix.  Make sure all learning rates, batch sizes, etc are reported]}
This appendix contains full details required to reproduce the image generation experiments that we reported in the main text.
Details for the standard \ac{dm} are contained in \Cref{subsec: standard DM implement MNIST}, while details for \ac{oism} are contained in \Cref{subsec: oism detail MNIST}.

\paragraph{Forward process and eigenfunctions} The \ac{ou} forward process takes the following form
\begin{align}
    \mathrm{d}\mbX_\tau = -\frac{\beta_\tau}{2} \mbX_\tau\mathrm{d}\tau + \sqrt{\beta_\tau} \mathrm{d}\mbW_\tau, \qquad \beta_\tau = \beta_0 + \tau(\beta_1-\beta_0), \qquad \mbX_0 \sim \rho_0,
\end{align}
where $\beta_0 = 0.1, \beta_1=20$. The above forward process links to the original \ac{ou} forward process \eqref{eq: OU process} by the change of time variable formula $t(\tau)=\frac{\beta_0\tau}{2} + \frac{(\beta_1-\beta_0)\tau^2}{4}$. 

\paragraph{Computational resources} 
We made use of a JAX implementation that runs on a NVIDIA A100 GPU. %\textcolor{red}{[ZS: remove for anonymity. Apart from A100, is there anything else to be mentioned?] }\mr{Perhaps the number of cards, memory of each card, and the storage needed}

\paragraph{Implementation details} 
%In addition to the model design choices \mr{what do you refer to here? } described in \Cref{app: experiment}
As \ac{oism} for image generation requires Hermite polynomials, we mainly use the Hermite recurrence relations mentioned in \Cref{app: hermite} to calculate quantities related to \ac{oism}. For the neural network code implementation, we adapt the JAX implementation of the score-based diffusion model \citep{songMaximumLikelihoodTraining2021} that was downloaded under the MIT license from \url{https://github.com/yang-song/score_flow}. \ac{dsm} \citep{hoDenoisingDiffusionProbabilistic2020} minimizes the following objective equivalent to the score matching loss \eqref{eq: score matching p0}
%\mr{perhaps put variables that we integrate in the expectation on the same line}
%The implementation is slightly modified for \ac{oism}.
% and we use differential equation solvers obtained from \url{https://github.com/rtqichen/torchdiffeq} to add an implementation of the probability flow \ac{ode} \eqref{eq: prob flow ode}. 
\begin{align}
    \int_0^1 w(\tau) \mbox{\large{$\mathbb{E}$}}_{\substack{\mbX_0\sim\rho_0,\\ \mbepsilon\sim\cN(\mathbf{0}, \mbI)}
    }\left[\big\lVert\mbepsilon-\widehat{\mbepsilon}_\tau(\alpha_\tau\mbx_0+\sigma_\tau\mbepsilon)\big\rVert^2\right]\mathrm{d}\tau, \label{eq: dsm}
\end{align}
given a positive-valued \emph{weighting} function $w(t)>0$, and $\alpha_\tau$ and $\sigma_\tau$ correspond to the noise perturbation level at time $\tau$. 
%\mr{shall we change the name of the weights to $w$, for consistency with the main text? Or maybe not, if you want to specify that the perturbation weight $\lambda$ cannot be blindly  applied  as weight for ISM etc}\mr{might be helpful to define here $\alpha_\tau, \sigma_\tau, \widehat{\mbepsilon}_\tau$}
\citet{songMaximumLikelihoodTraining2021} discover that minimizing \eqref{eq: dsm} with $w(\tau) = \beta_\tau/\sigma_\tau^2$ is equivalent to maximizing $\mathbb{E}_{\mbX_0\sim\rho_0} \left[\log \rho_0^{(\text{SDE})}(\mbX_0)\right]$, where $\rho_0^{(\text{SDE})}$ represents the marginal distribution of the \ac{sde}-based sampler driven by $\widehat{\mbepsilon}_\tau$. We follow the same likelihood weighting function and the variance reduction techniques proposed by \citet{songMaximumLikelihoodTraining2021}.
%\mr{perhaps add a line to say what the importance sampling achieves - variance reduction}. 

In principle, neural networks that optimize loss functions equivalent up to a constant (such as \ac{dsm} and \ac{ism}), or whose optimal point can be recovered from these (as in \ac{oism})
should have similar performance. However, because optimization is performed with respect to different variables (the random perturbation vector in \ac{dsm}, the score in \ac{ism}, or yet the gradient of the log-density ratio in \ac{oism}), one is required to designate what target function the neural network is attempting to estimate. As the neural network estimation target changes, the training process requires different hyperparameters; see Table 1 of \citet{salimans2022progressive}.
%cannot expect that the same weighting function carries the same meaning or performance  
% It is important to notice that in \ac{dsm} the output of a neural network can be interpreted as an approximation of the random perturbation vector $\mbepsilon$, the score function $\nabla\log\rho_t$ or the gradient of the log density-ratio $\nabla\log\rho_t/\pi$, they do not interact with the same $\lambda(t)$ weighting function in the same way; 
%\textcolor{teal}{ZS: see Table 1 of \citet{salimans2022progressive}. this is a good reference}
In our scenario, we opt for the maximum likelihood weighting function, and choose the neural network output as an approximation of the perturbation vector $\mbepsilon$. 
\paragraph{Dataset}
The CIFAR-10 dataset was obtained via the TensorFlow Datasets API \url{https://github.com/tensorflow/datasets}, which fetches the data from its original source at \url{https://www.cs.toronto.edu/~kriz/cifar.html} under the terms of the MIT License. 
% The MNIST dataset was downloaded from \url{http://yann.lecun.com/exdb/mnist/} under the terms of the Creative Commons Attribution-Share Alike 3.0 license. 
%The image pixel values were normalized to the range  $[-1, 1]$ for the purpose of constructing a \ac{dm}. \mr{can you explain this better?}
While the original CIFAR-10 images have integer pixel values ranging from $0$ to $255$, we normalize them to between $-1$ and~$1$. 
% \textcolor{blue}{[Update to CIFAR-10]}
%

\subsection{Standard Denoising Diffusion Probabilistic Model}
\label{subsec: standard DM implement MNIST}
For the reasons explained above, we plug in the neural network output $\texttt{NN}_\mbtheta(\mbx, \tau)$ as the $\widehat{\mbepsilon}_\tau(\mbx)$ for the \ac{dsm} objective function \eqref{eq: dsm}.

%We opt for the interpretation of the neural network output $\texttt{NN}_\mbtheta(\mbx, \tau)$ as the approximation of the noise perturbation vector, i.e., 
%\begin{align*}
%    \widehat{\mbepsilon}_\tau(\mbx) = \texttt{NN}_\mbtheta(\mbx, \tau).
%\end{align*}
Following common experimental practice for image generation tasks, we use a UNet architecture for score estimation. Specifically, we adopted a UNet consisting of 4 ResNet blocks and learned time embeddings\footnote{Full details are provided as part of the code contained in the electronic supplement.}. 
A total of 500,000 iterations of Adam were performed, with batch size of 128, and the UNet parameters were optimized with learning rate $2 \times 10^{-4}$. 
The total running time was about 25 hours.% \mr{in the main paper we said 30, below 27}
\subsection{Operator-Informed Score Matching} 
\label{subsec: oism detail MNIST}
\ac{oism} gives $\tilde{\mbs}_t(\mbx)$ \eqref{eq: oism final}, an estimate of $\nabla\log\rho_t/\pi(\mbx)$. However, 
% In order to combine the \ac{oism} output with that of the standard \ac{dm} \Cref{subsec: standard DM implement MNIST}, we implement the transformation  
we are interested using the same approximation type as described in \Cref{subsec: standard DM implement MNIST} for maximum comparability. Therefore, 
%\textcolor{red}{\sout{ we construct} 
in order to train a neural network on the residual of the \ac{oism} score, we construct 
the following
%When combining \ac{oism} with the output of a neural network, we still require an approximation of the perturbation vector. Given an \ac{oism} approximation $\tilde{\mbs}_t(\mbx)$ of $\nabla\log\rho_t/\pi(\mbx)$, we obtain the following combined estimator
\begin{align*}
    \widehat{\mbepsilon}_\tau(\mbx) = \sigma_\tau\left(\mbx-\tilde{\mbs}_{t(\tau)}(\mbx)\right) + \texttt{NN}_{\mbtheta}(\mbx, \tau).
\end{align*}

The Hermite polynomials are implemented using the recurrence relations \eqref{eq: hermite recurrence}, and the interaction coefficients are calculated using the dynamic programming rule \eqref{eq: hermite interaction relation}. Unlike \Cref{app: experiment}, we do not apply shrinkage estimators but instead evaluate  $\mathbb{E}_{\rho_0}[\phi_k]$ using sample means; the linear systems for finding the optimal \ac{oism} coefficients are solved using a preconditioner as per \Cref{app: preconditioner}.

\begin{figure}[t!]
    \centering
    \includegraphics[width=0.9\linewidth]{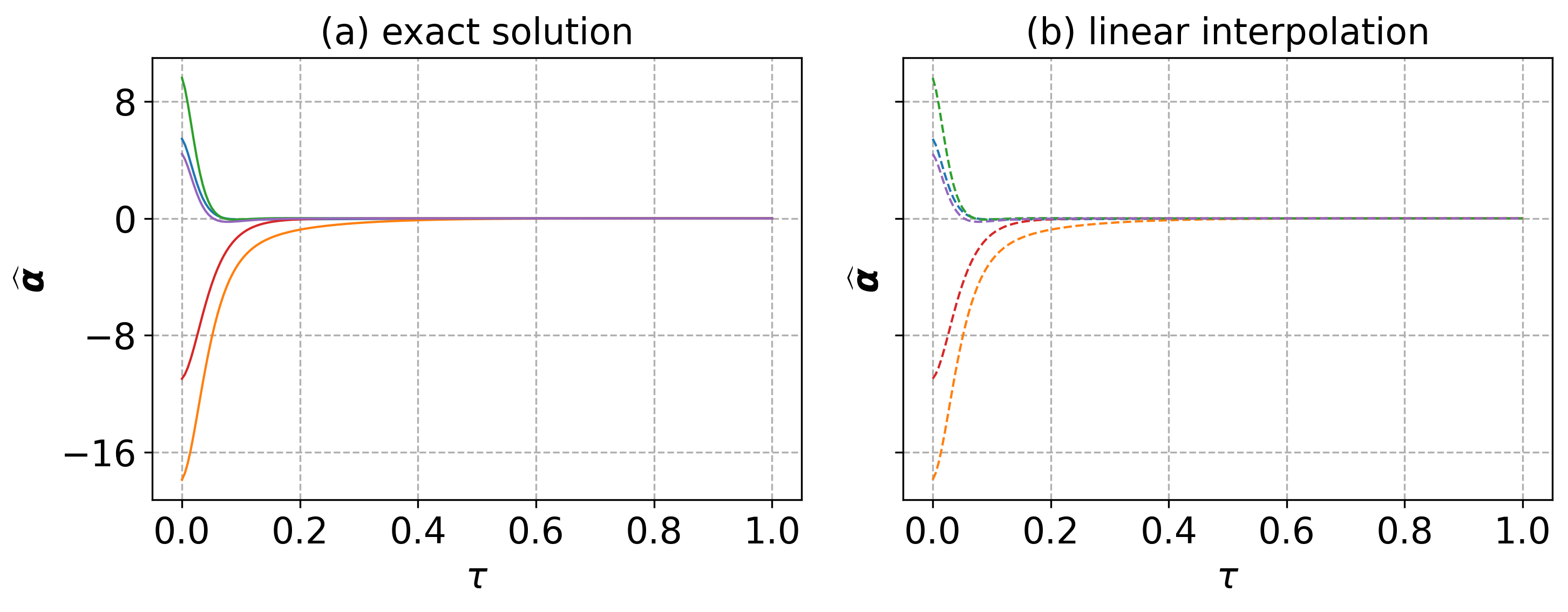}
    \caption{Coefficients $\widehat{\boldsymbol{\alpha}}$ from \textbf{OISM} (6) as a function of the normalized time $\tau$. Left: exact solution on the grid of points at which the linear systems were solved during training. Right: linearly interpolated solutions obtained on a grid of 1,000 evenly-spaced time points. 
    Simple linear interpolation is sufficient for training and evaluation of the neural network, due to the temporal correlation.} 
    \label{fig: temporal oism}
\end{figure}

\paragraph{Pre-solving linear systems and linear interpolation} The \ac{oism} score estimates are fixed throughout the training and evaluation processes, and the sequence of linear systems $\left(\mbA_t^{-1}\mbb_t\right)_{t\geq 0}$ is correlated temporally, as illustrated in \Cref{fig: temporal oism}.
For both training and evaluation, we obviate the need to call a linear solver for every score evaluation by a pre-processing step, that is, we
%We choose not to call a linear solver for every iteration \textcolor{red}{of the optimizer, what about evaluation?} by 
pre-solve the equations $\mbA_\tau\widehat{\mbalpha}=\mbb_\tau$ on an evenly-spaced grid of 1,000 $\tau$s between $0$ and $1$, and extend the solution to continuous times by linear interpolation. 

\begin{table}[t]
  \centering
  %\footnotesize
  %\setlength{\tabcolsep}{8pt}
    \caption{Average runtime for each training step (in seconds). Interpolating the solutions of the linear system introduces a small computational overhead.}
  \begin{tabular}{lcc|c}
    \toprule
    Method          & OISM (3)  & OISM (6) & DDPM \\
    \midrule
    Exact solution         & 0.428    & 0.799   & \multirow{2}{*}{0.308} \\
    Linear interpolation   & 0.315    & 0.312   &                        \\
    \bottomrule
  \end{tabular}
  \label{tab: runtime}
\end{table}

%Here we provide some brief remarks on \ac{oism} that are necessary to understand how our experiments can be reproduced.
%Recall that the linear score approximation $\hat{\mbs}_t(\mbx)$ provided by \ac{oism} requires that the matrices $\alpha_t^2\widehat{\mbSigma} + \sigma_t^2\mbI$ in \eqref{eq: lin estimator} be inverted.
%This was achieved by first obtaining an eigendecomposition $\mbV\text{diag}(\mbpsi)\mbV^\top$ of $\widehat{\mbSigma}$, yielding 
%\begin{align*}
%    \left(\alpha_t^2\widehat{\mbSigma} + \sigma_t^2\mbI\right)^{-1} = \mbV \text{\diag}\left(\frac{1}{\alpha_t^2\psi_d + \sigma_t^2}\right)\mbV^\top.
%\end{align*}
%Similar to the toy experiments that we reported, here we used the UNet architecture as a flexible model for the residual $\mbr_t(\mbx)$, as described in the main text. 

%For this experiment we ran a total of 50,0000 iterations with batch size of 128 for fair comparison, and UNet parameters were optimized using Adam with learning rate $2\times 10^{-4}$ as well.
% -- the extra linear term of \ac{oism} allowed for a higher learning rate in this experiment. 
The training time via pre-solving and interpolation was similar to that of a neural network only DDPM in \Cref{subsec: standard DM implement MNIST}. 
%around 27 hours \textcolor{red}{\sout{-- it took negligibly longer}, this is slightly more } than the standard DDPM \mr{here you mean \ac{dsm} or the OISM and NN but no pre-solving?} 
The extra computation required for interpolating the \ac{oism} solutions constituted a small fraction of the overall computational cost; see \Cref{tab: runtime}. 